\documentclass[10pt,twocolumn,letterpaper]{article}

\usepackage{cvpr}
\usepackage{times}
\usepackage{epsfig}
\usepackage{graphicx}
\usepackage{amsmath}
\usepackage{amssymb}

\usepackage[utf8]{inputenc} 
\usepackage[T1]{fontenc}    
\usepackage{hyperref}       
\usepackage{url}            
\usepackage{booktabs}       
\usepackage{amsfonts}       
\usepackage{nicefrac}       
\usepackage{microtype}      
\usepackage{booktabs} 
\usepackage{multirow}
\usepackage{subfigure}
\usepackage{graphicx}
\usepackage{float}
\usepackage{diagbox}
\usepackage{xcolor}
\usepackage{verbatim}
\newcommand{\vsss}{vs.}

\usepackage[pagebackref=true,breaklinks=true,letterpaper=true,colorlinks,bookmarks=false]{}

 \cvprfinalcopy 

\def\cvprPaperID{6546} 

\ifcvprfinal\fi
\begin{document}


\title{Global Texture Enhancement for Fake Face
Detection in the Wild}


\author{Zhengzhe Liu, Xiaojuan Qi$^{1,2}$,  Philip H. S. Torr$^1$ \\
$^1$University of Oxford, $^2$The University of Hong Kong\\
}

%

\newcommand{\fix}{\marginpar{FIX}}
\newcommand{\new}{\marginpar{NEW}}

\newcommand{\tabincell}[2]{\begin{tabular}{@{}#1@{}}#2\end{tabular}}






\maketitle
\footnotetext[1]{The first three are real and the last three are fake.}
\begin{abstract}
Generative Adversarial Networks (GANs) can generate realistic fake face images that can easily fool human beings. On the contrary, a common Convolutional Neural Network (CNN) discriminator can achieve more than $99.9\%$ accuracy in discerning fake/real images. In this paper, we conduct an empirical study on fake/real faces, and have two important observations: firstly, the texture of fake faces is substantially different from real ones; secondly, global texture statistics are more robust to image editing and transferable to fake faces from different GANs and datasets. Motivated by the above observations, we propose a new architecture coined as Gram-Net, which leverages global image texture representations for robust fake image detection. Experimental results on several datasets demonstrate that our Gram-Net outperforms existing approaches. Especially, our Gram-Net is more robust to image editings, e.g. down-sampling, JPEG compression, blur, and noise. More importantly, our Gram-Net generalizes significantly better in detecting fake faces from GAN models not seen in the training phase and can perform decently in detecting fake natural images.
\end{abstract}

\vspace{-0.2cm}
\section{Introduction}
\begin{figure}
 \centering
  \begin{tabular}{@{\hspace{0.1mm}}c@{\hspace{0.1mm}}c@{\hspace{0.1mm}}c}
  \includegraphics[width=0.3\linewidth]{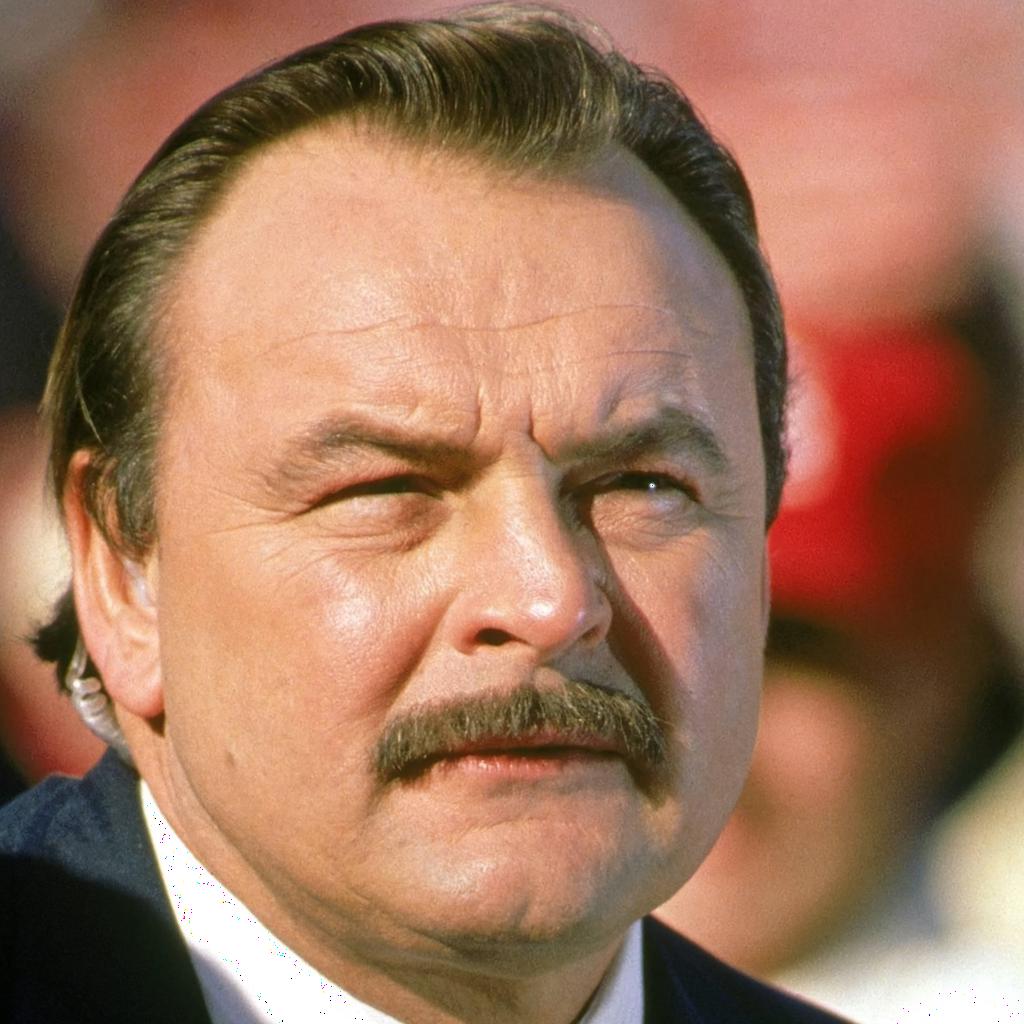} &
    \includegraphics[width=0.3\linewidth]{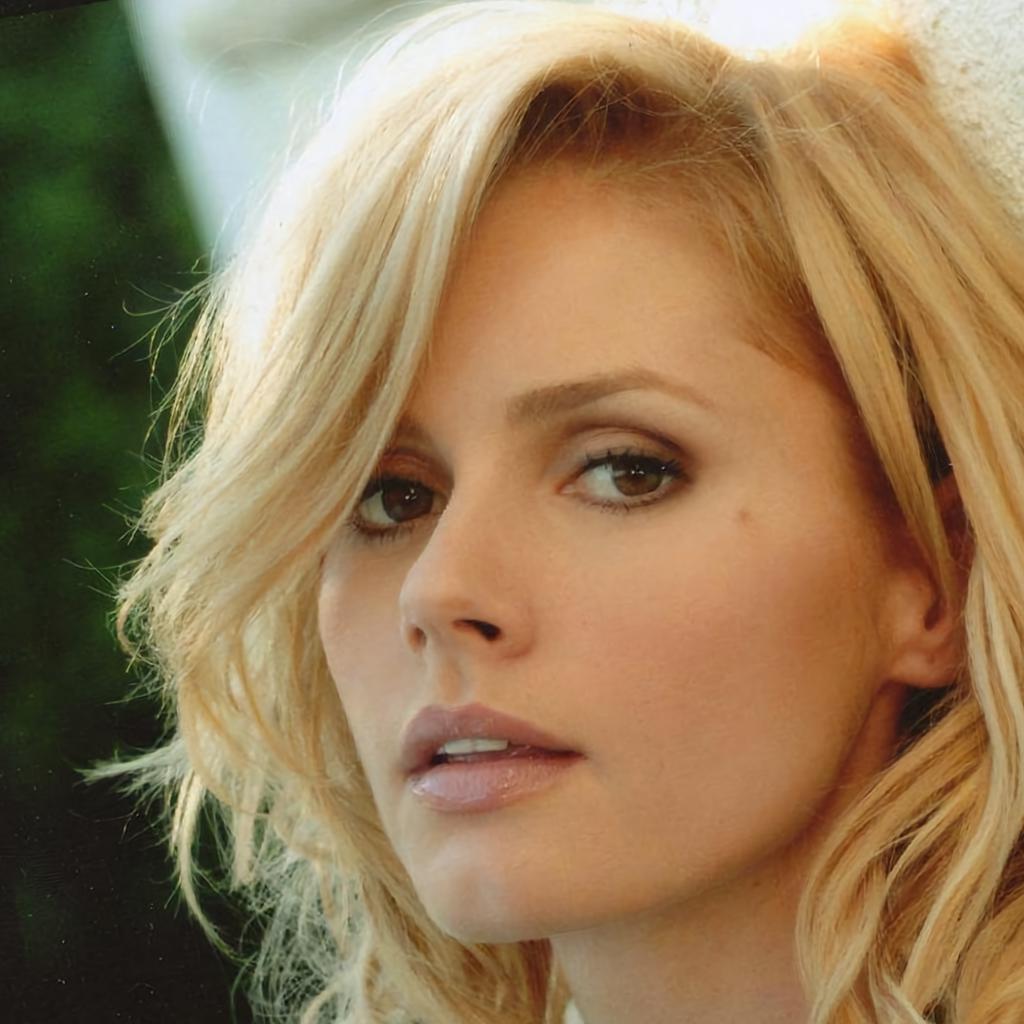}&
    \includegraphics[width=0.3\linewidth]{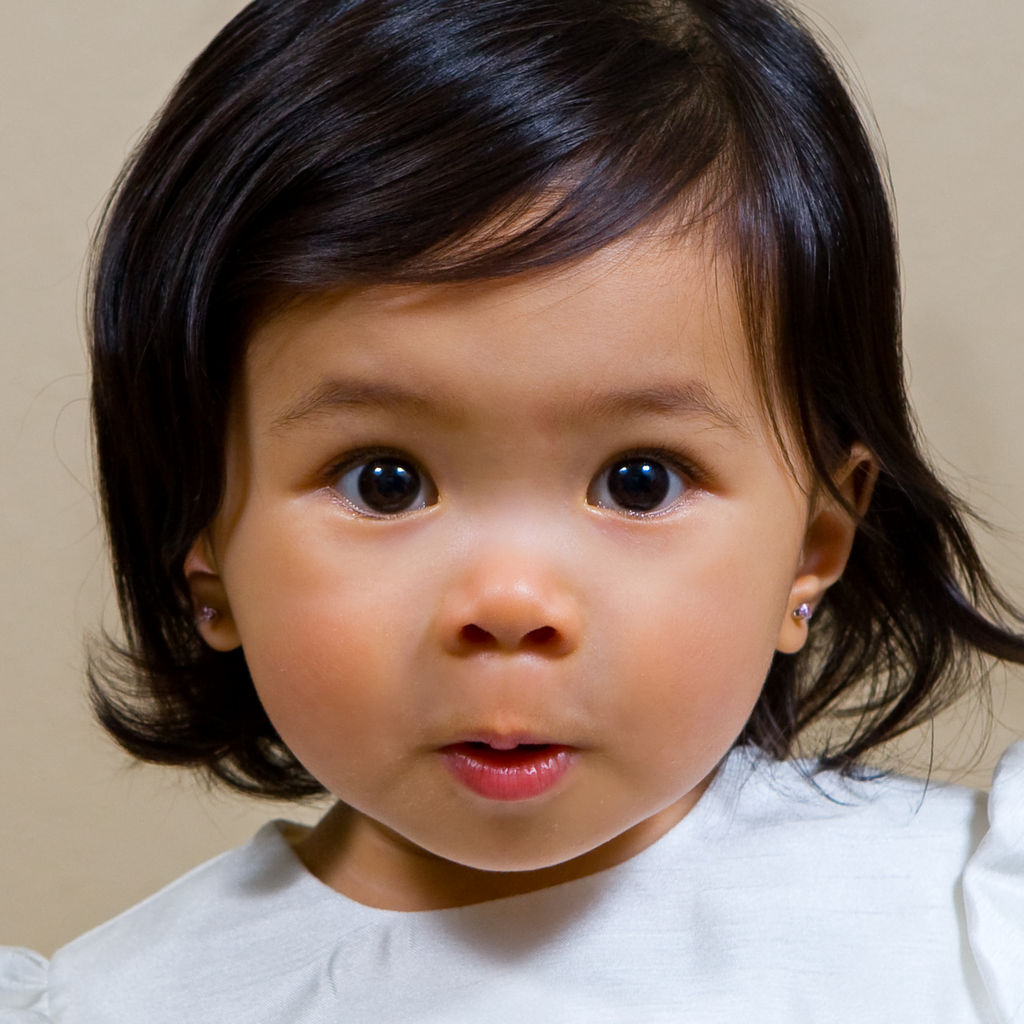}\\
    \includegraphics[width=0.3\linewidth]{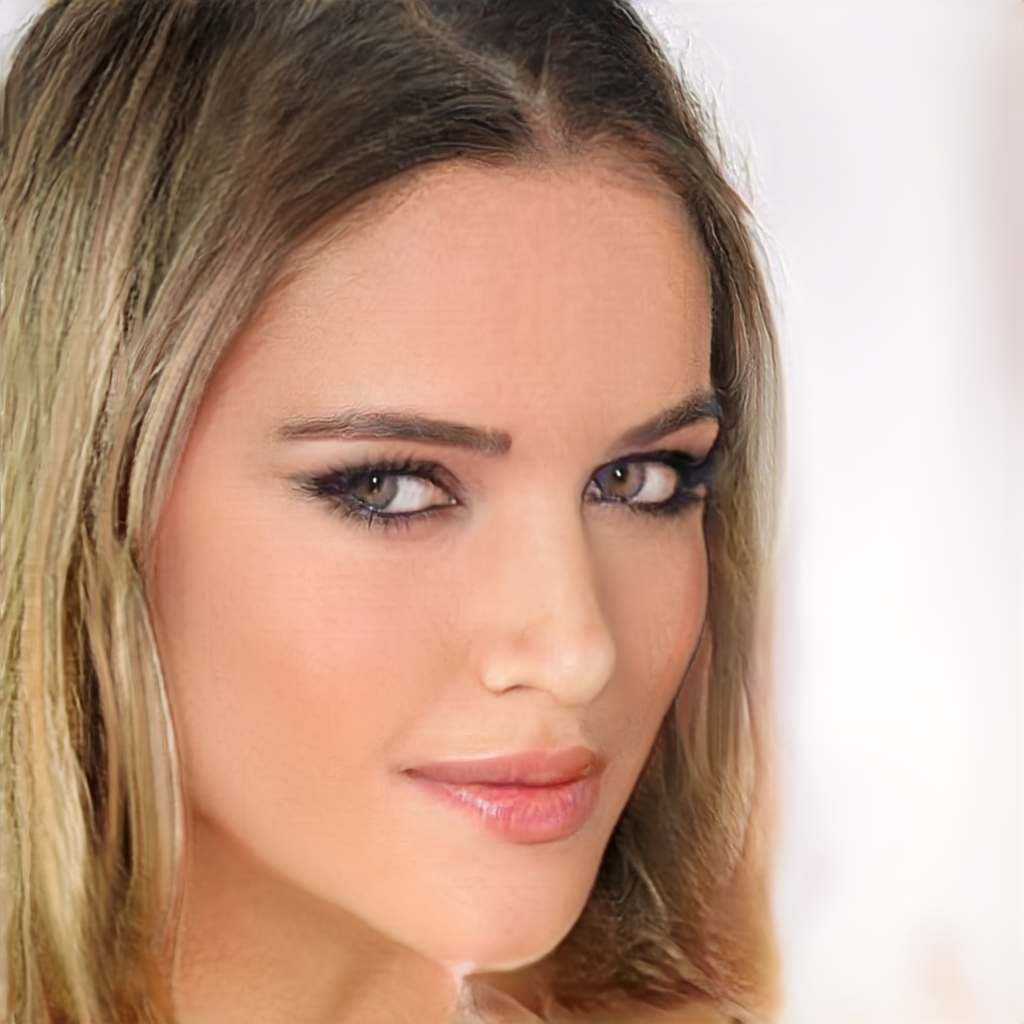}&
    \includegraphics[width=0.3\linewidth]{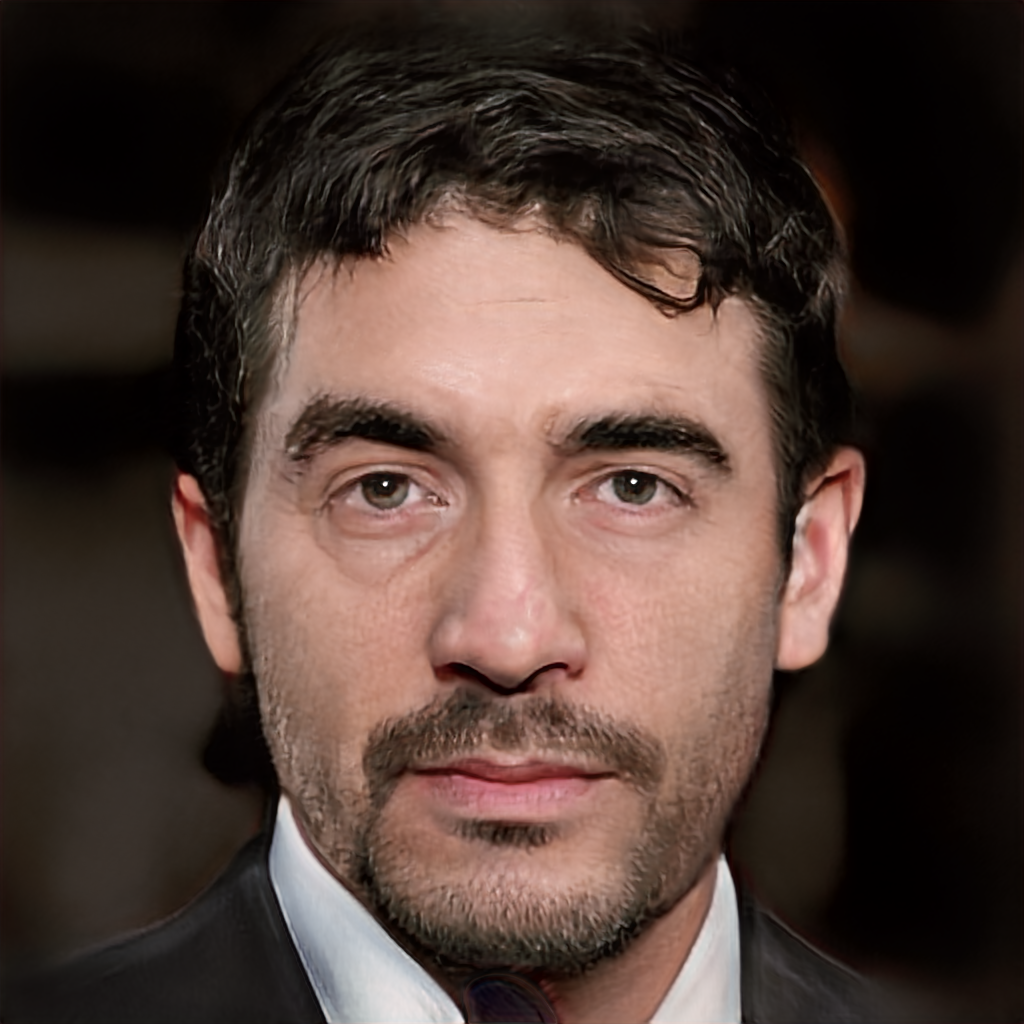}&
    \includegraphics[width=0.3\linewidth]{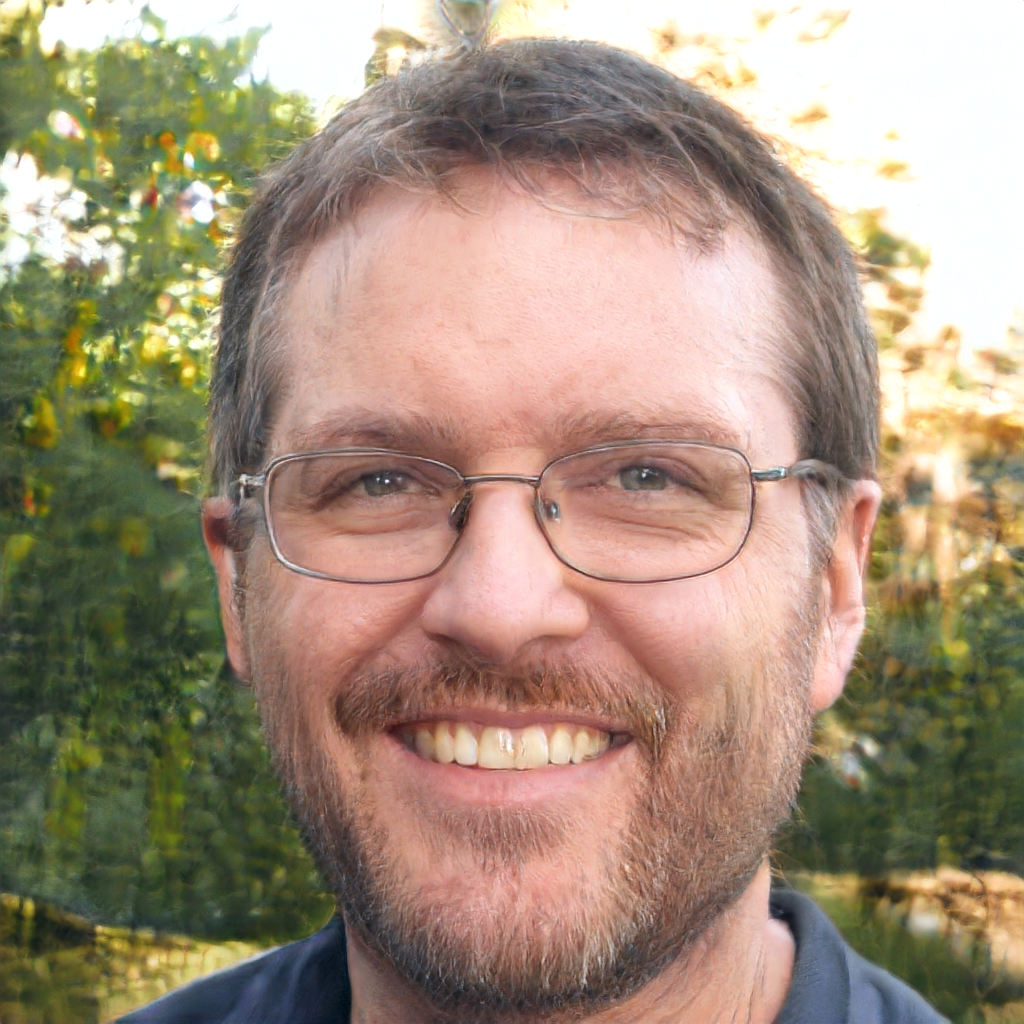}
    \end{tabular}
  \caption{Can you determine which are real and which are fake? (answer key below)$^1$}      
  \label{fig:ims} 
\end{figure}

With the development of GANs ~\cite{gulrajani2017improved,karras2017progressive,karras2018style,arjovsky2017wasserstein}, computers can generate vivid face images that can easily deceive human beings as shown in Figure~\ref{fig:ims}. (Can you guess which images are generated from GANs?)
These generated fake faces will inevitably bring serious social risks, \eg fake news and evidence, 
and pose threats to security.
Thus, powerful techniques to detect these fake faces are highly desirable.
However, in contrast to the intensive studies in GANs, our understanding of generated faces is fairly superficial and how to detect fake faces is still an under-explored problem.
Moreover, fake faces in practical scenarios are from different unknown sources, {\ie} different GANs, and may undergo unknown image distortions such as downsampling, blur, noise and JPEG compression, which makes this task even more challenging.
In this paper, we aim to produce new insights on understanding fake faces from GANs and propose a new architecture to tackle the above challenges. Our contributions are as follows.
\vspace{-0.2cm}
\paragraph{Contribution 1.} To facilitate the understanding of face images from GANs, we systematically study the behavior of human beings and CNN models in discriminating fake/real faces detailed in Section~\ref{sec:human}. In addition, we conduct extensive ablation experiments to diagnose the CNN discriminator and perform low-level statistics analysis as verification.

These empirical studies lead us to the following findings.
\begin{itemize}
    \item \textit{Texture statistics} of fake faces are substantially different from natural faces.
    \item Human focus on visible shape/color artifacts to detect fake face while CNNs focus more on \textit{texture} regions. 
    \item CNNs take \textit{textures} as an important cue for fake face detection.
    A ResNet model performs almost perfectly in detecting untouched fake faces if the training data and testing data are from the same source.
\end{itemize}
\vspace{-0.3cm}
\paragraph{Contribution 2.} Although a CNN based fake face detector performs significantly better than human beings, it is still not robust enough to handle real-world scenarios, where images may be modified and/or from different unknown sources.
With further analysis of the relationship between \textit{texture} and fake face detection, we found \textit{large texture} information is more robust to image distortions and more invariant for face images from different GANs.
However, CNNs cannot fully capture \textit{long-range or global} cues due to their limited effective receptive field as studied in~\cite{luo2016understanding}.
Motivated by the above observation, we further develop a novel architecture -- Gram-Net, which improves the robustness and generalization ability of CNNs in detecting fake faces. 
The model incorporates ``Gram Block'' into the CNN backbone shown in Figure \ref{fig:gramnet}.
The introduced Gram layer computes \textit{global texture} representations in multiple semantic levels, which complements the backbone CNN.
\vspace{-0.2cm}
\paragraph{Contribution 3}  Experiments on fake faces from StyleGAN \cite{karras2018style}, PGGAN \cite{karras2017progressive}, DRAGAN \cite{kodali2017convergence}, DCGAN \cite{radford2015unsupervised},  StarGAN \cite{choi2018stargan}, and real faces from CelebA-HQ \cite{karras2017progressive}, FFHQ \cite{karras2018style}, CelebA \cite{liu2015faceattributes}, show that our Gram-Net achieves state-of-the-art performance on fake face detection. Specifically, our proposed Gram-Net is robust for detecting fake faces which are edited by resizing ($10\%$ improvement), blurring ($15\%$ improvement), adding noise ($13\%$ improvement) and JPEG compressing ($9\%$ improvement).
More importantly, Gram-Net demonstrates significantly better generalization abilities.
It surpasses the compared approaches by a large margin (more than $10\%$ improvement) to detect fake faces generated by GANs that are not seen in the training phase and GANs trained for other tasks including image-to-image translation GANs, {\eg} StarGAN.
Further, our experiments show that Gram-Net (trained on StyleGAN) generalizes much better with a $10\%$ improvement to detect fake natural images from GANs trained on ImageNet \cite{krizhevsky2012imagenet}, {\eg} BigGAN ~\cite{brock2018large}.

\section{Related work}
\vspace{0.1cm}
\paragraph{GANs for human face generation.}
Recently, GAN models ~\cite{goodfellow2014generative,radford2015unsupervised,kodali2017convergence,arjovsky2017wasserstein,berthelot2017began,karras2017progressive,karras2018style,liu2017unsupervised,zhu2017unpaired,choi2018stargan} have been actively studied with applications for face image generation.
One stream of research is to design GANs ~\cite{goodfellow2014generative,radford2015unsupervised,kodali2017convergence,arjovsky2017wasserstein,berthelot2017began} for generating random face images from random vectors.
Early works ~\cite{goodfellow2014generative,radford2015unsupervised,kodali2017convergence,arjovsky2017wasserstein,berthelot2017began} can generate high quality low resolution images but suffer from mode collapse issues for generating high resolution images.
The most advanced high resolution ($1024 \times 1024$) GAN models -- PGGAN~\cite{karras2017progressive} and StyleGAN~\cite{karras2018style}-- can generate high quality face images that can even fool human beings.
Another stream is to utilize GAN models for image-to-image translation tasks~\cite{liu2017unsupervised,zhu2017unpaired,choi2018stargan}, {\eg}, Choi {\etal} proposed  StarGAN model which can perform face image to face image translation.
These generated fake faces may cause negative social impact. 
In this work, we aim to help the community gain more understanding about GAN generated fake faces and introduce novel neural network architecture for robust fake face image detection.
\paragraph{Fake GAN face detection.}
Recently, some researchers have investigated the problem of fake face detection~\cite{li2018detection,mccloskey2018detecting,nataraj2019detecting,marra2018detection,marra2019gans,xuan2019generalization,zhang2019detecting,wang2019fakespotter}. Color information is exploited in~\cite{li2018detection,mccloskey2018detecting}.
In contrast, we found the performance of the CNN models changes little even if color information is removed.
Marra \etal~\cite{marra2018detection} showed that each GAN leaves specific finger-prints on images, and proposed to identify the source generating these images.
However, the method cannot generalize to detect fake faces from GAN models that do not exist in the training data.
Xuan \etal~\cite{xuan2019generalization} adopted data augmentation for improving generalization, nevertheless, further improvements are limited by the detection algorithm.
Nataraj \etal~\cite{nataraj2019detecting} proposed to take a color co-occurrence matrix {as input} for fake face detection. {However, the hand-craft feature input results in losing the information of raw data}.
Zhang \etal~\cite{zhang2019detecting} designed a model to capture the artifacts caused by the decoder.
However, it failed to detect fake images from GANs with drastically different decoder architecture which is not seen in the training phase, while our approach can handle this case effectively.
Wang \etal~\cite{wang2019fakespotter} proposed a neuron coverage based fake detector. However, the algorithm is time-consuming, hard to be deployed in real systems, and the performance is still far from satisfactory.
Marra \etal~\cite{marra2019incremental} detected fake images with incremental learning. However, it only works when many GAN models are accessible in the training phase.
Other works \cite{li2018exposing,yang2019exposing} focused on the alignment of face landmarks to check whether the face is edited by face-swapping tools like DeepFakes~\cite{liu2017unsupervised}.
Unlike the above, we intensively analyze fake faces, and correspondingly propose a novel simple framework which is more robust and exhibits significantly better generalization abilities.

\vspace{-0.6cm}

\paragraph{Textures in CNNs.}
The texture response of CNNs has attracted increasing attention in the last few years.
Geirho \etal~\cite{geirhos2018imagenet} showed that CNN models are strongly biased on textures rather than shapes.
Our empirical study also reveals that CNN can utilize texture for fake face detection which is in line with the findings in~\cite{geirhos2018imagenet}. Motivated by the above observation, we further analyzed texture differences in terms of low-level statistics.
Gatys~\etal \cite{gatys2015texture} proposed that the Gram matrix is a good description of texture, which is further utilized for texture synthesis and image style transfer~\cite{gatys2016image}.
The above works exploit the Gram matrix for generating new images by constructing Gram matrix based matching losses.
Our work is related to these methods by resorting to the Gram matrix.
However, different from \cite{gatys2016image,gatys2015texture},
our work adopts the Gram matrix as a global texture descriptor to improve discriminative models and demonstrates its effectiveness in improving robustness and generalization.




\section{Empirical Studies and Analysis}



\subsection{Human {\vsss} CNN}\label{sec:human}

\begin{figure*}
  \centering
  \subfigure[\footnotesize{Real}]{
    \label{fig:subfig:a} 
    \includegraphics[width=1.1in]{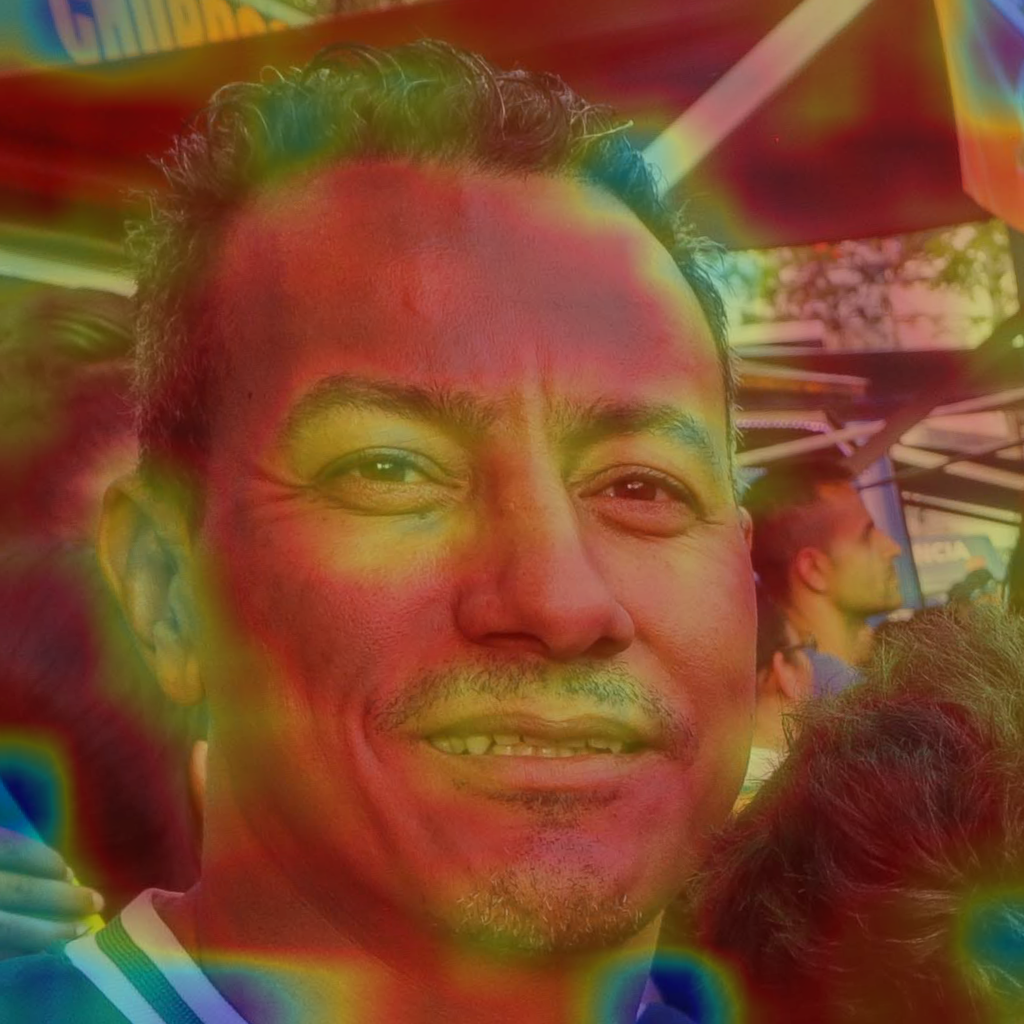}}
  \subfigure[Real]{
    \label{fig:subfig:b} 
    \includegraphics[width=1.1in]{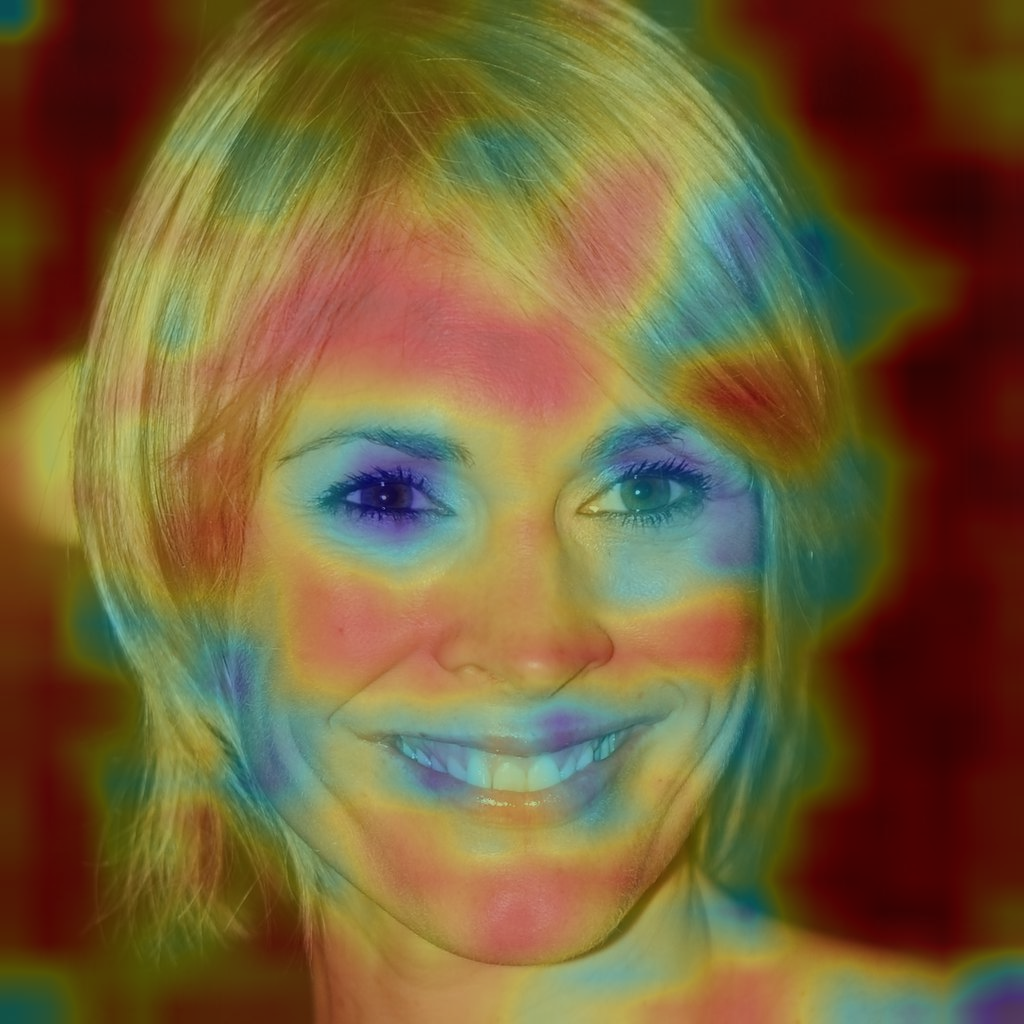}}
 \subfigure[Fake]{
    \label{fig:subfig:b} 
    \includegraphics[width=1.1in]{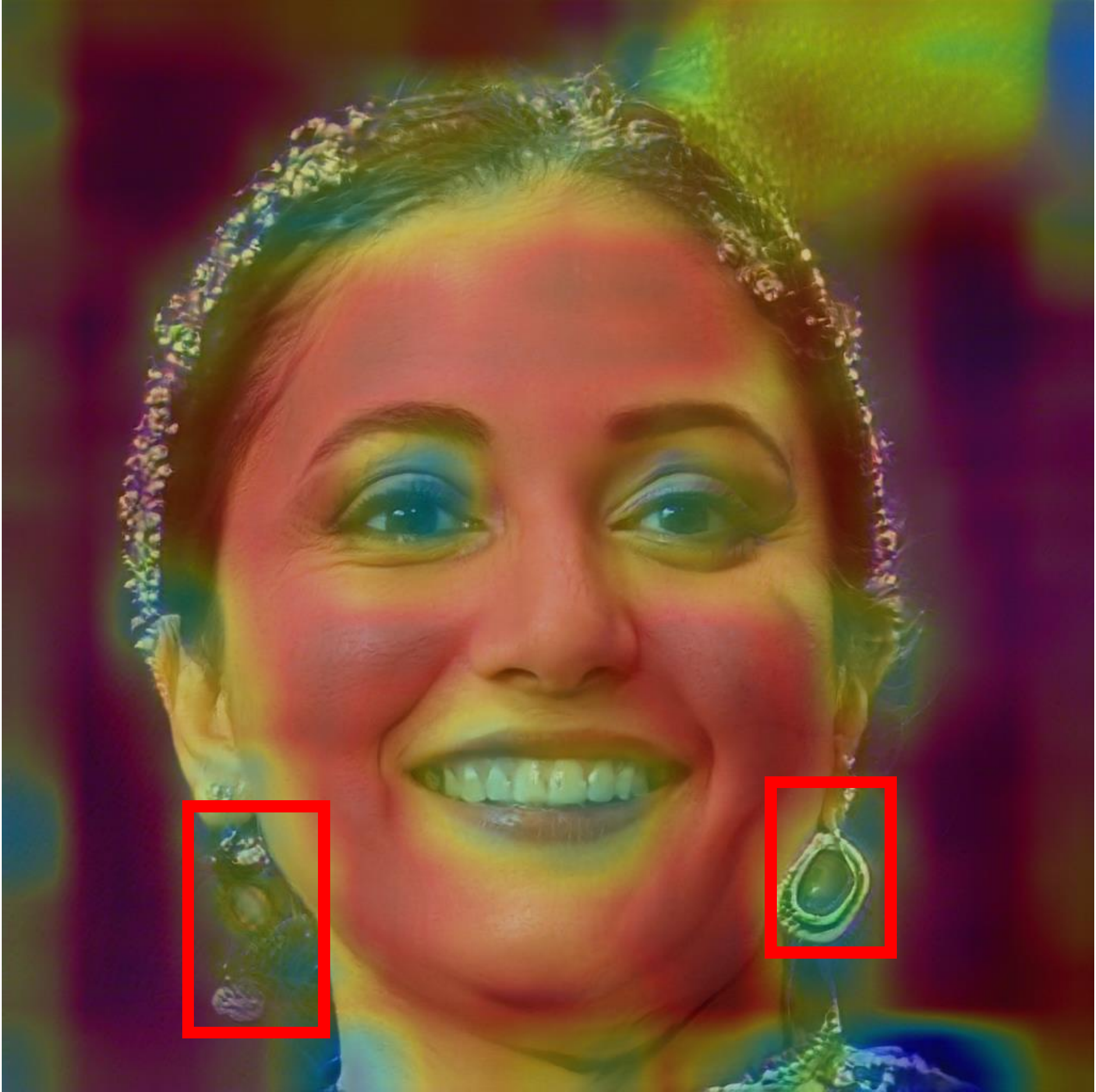}}
  \subfigure[Fake]{
    \label{fig:subfig:b} 
    \includegraphics[width=1.1in]{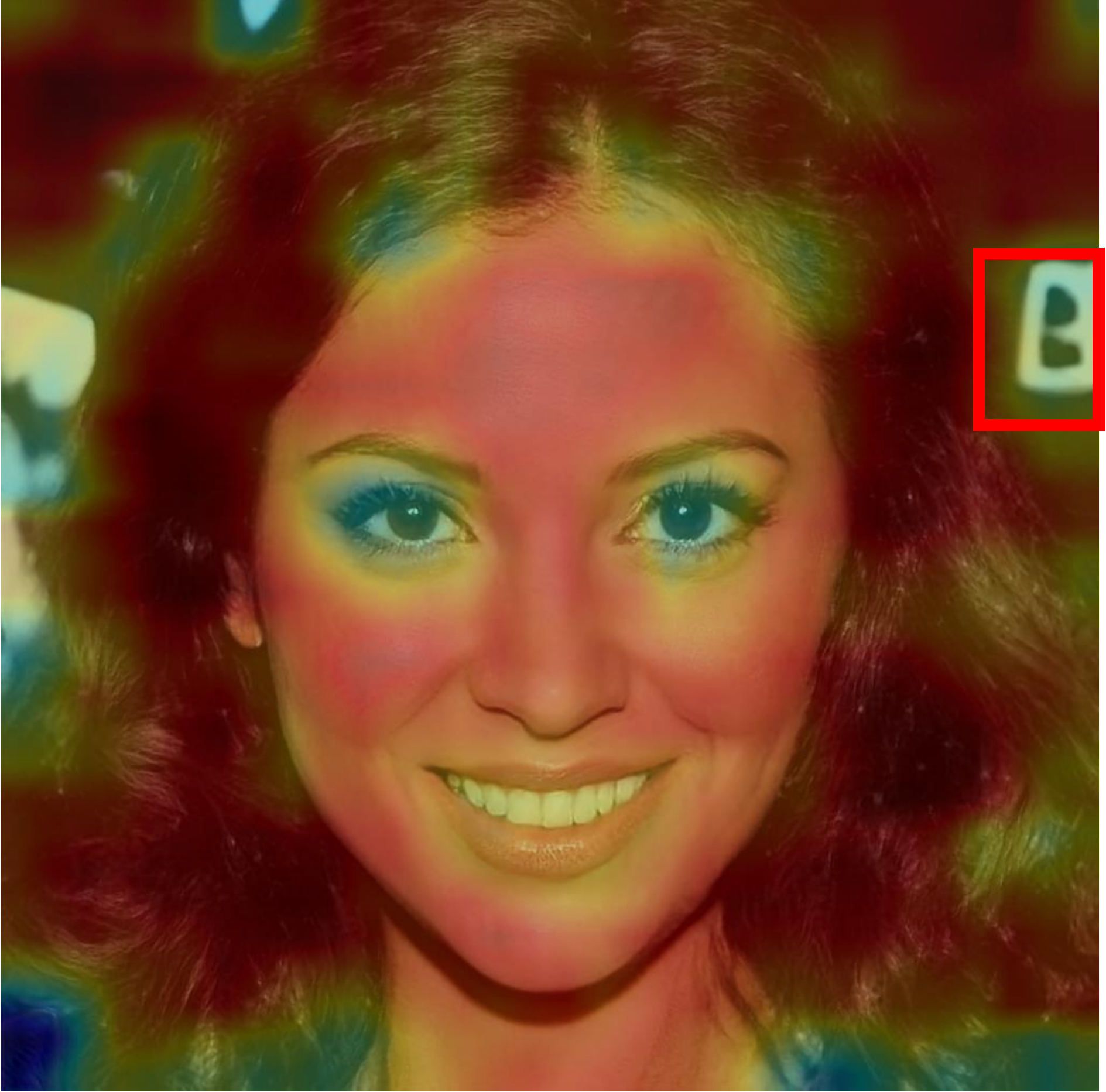}}
  \subfigure[Fake]{
    \label{fig:subfig:b} 
    \includegraphics[width=1.1in]{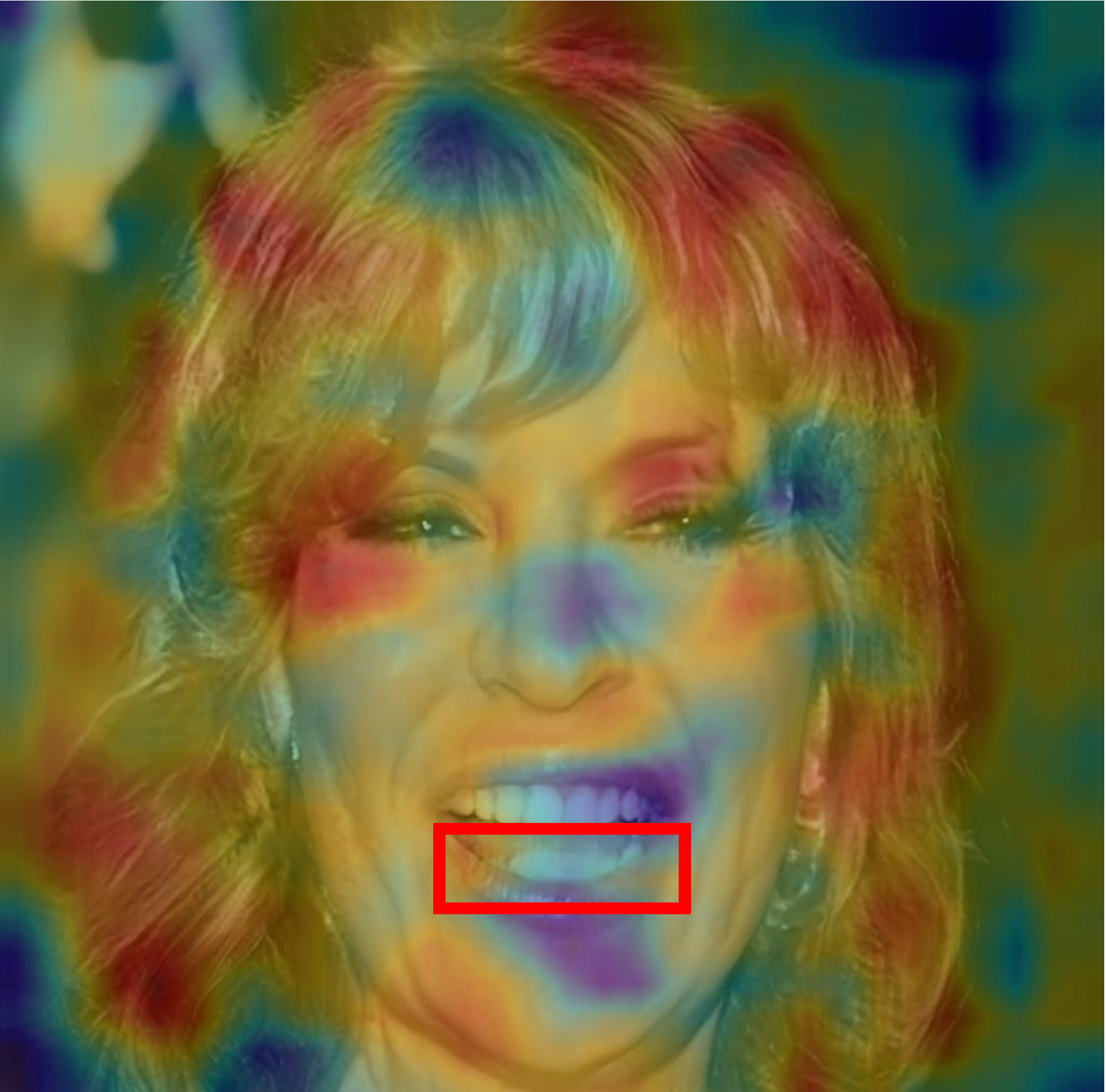}}

  \caption{Class activation maps from trained ResNet model (better viewed in color). 
  The red bounding box shows the visible artifacts indicated by human observers but activated weakly by CNN: (c) asymmetrical earrings; (d) irregular letter; (e) irregular teeth.}
  \label{fig:attention} 
\end{figure*}

To shed insights on understanding fake faces generated form GANs, we systematically analyze the behavior of human beings and CNNs in discerning fake/real faces by conducting psychophysical experiments.
Specifically, our experiments are performed in \textit{in-domain} setting, where the model is trained and tested on fake images from the same GAN.
\vspace{-0.2cm}
\paragraph{User study.} For each participant, we firstly show him/her all the fake/real faces in the training set (10K real and 10K fake images).
Then a randomly picked face image in our test set is shown to him/her without a time limit.
Finally, he/she is required to click the ``real'' or ``fake'' button.
On average, it takes around $5.14$ seconds to evaluate one image. The results in this paper are based on a total of 20 participants, and each participant is required to rate 1000 images.
At the same time, we also collected the user's judgment basis if his/her selection was ``fake''. According to their votings, human users typically take as evidence easily recognized shape and color artifacts such as ``asymmetrical eyes'', ``irregular teeth'', ``irregular letters'', to name a few.
\vspace{-0.2cm}
\paragraph{CNN study and results.} Testing images are also evaluated by CNN model -- ResNet-18~\cite{he2016deep}. The training and testing follow the \textit{in-domain} setup.
Table \ref{tab:face} (row1 \& row2) shows that human beings are easily fooled by fake faces. 
In contrast, the ResNet CNN model achieves more than $99.9\%$ accuracy in all experiments.
\vspace{-0.2cm}
\paragraph{Analysis.}
To gain a deeper understanding about the question ``{Why CNNs perform so well at fake/real face discrimination?}'' and ``{What's the intrinsic difference between fake and real faces?}'', we further exploited CAM~\cite{zhou2016learning} to reveal the regions that CNNs utilize as evidence for fake face detection. Representative classification activation maps are shown in Figure~\ref{fig:attention}.
We can easily observe that the discriminative regions (warm color regions in Figure~\ref{fig:attention}) for CNNs mainly lie in the \textit{texture} regions, {\eg} skin and hair, while the regions with clear artifacts make little contribution (cold color, red bounding box in Figure~\ref{fig:attention}).
The above observation
motivates us to further study whether \textit{texture} is an important cue that CNNs utilize for fake face detection and whether fake faces are different from real ones regarding \textit{texture} statistics.

\begin{table*}
\centering
\scalebox{0.8}{
\begin{tabular}{c|c|c|c|c}
\toprule
Input & Human {\vsss} CNNs & {StyleGAN {\vsss} CelebA-HQ} & StyleGAN {\vsss} FFHQ & PGGAN {\vsss} CelebA-HQ   \\
\midrule
Full image& Human Beings & 75.15\% &63.90\% &79.13\% \\
Full image& ResNet &  99.99\% &99.96\% & 99.99\% \\
\midrule
Original (skin) & ResNet &99.93\%&99.61\%&99.96\%\\
Gray-scale (skin)& ResNet &  99.76\% &99.47\% & 99.94\%\\
L0-filtered (skin) & ResNet & 78.64\%& 76.84\% & 72.02\% \\

 \bottomrule
\end{tabular}
}
\vspace{0.3cm}
\caption{Quantitative results on fake face detection of human beings and CNNs, and skin region ablation studies in the \textit{in-domain} setting.}
\label{tab:face}
\end{table*}


\subsection{Is texture an important cue utilized by CNNs for fake face detection?}\label{sec:texture-important}

\begin{figure}
  \centering
  \subfigure[Real]{
    \label{fig:subfig:a} 
    \includegraphics[width=0.8in]{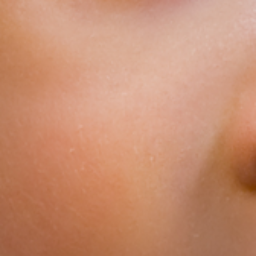}}
  \subfigure[Fake]{
    \label{fig:subfig:b} 
    \includegraphics[width=0.8in]{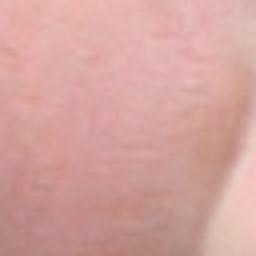}}
  \subfigure[Real]{
    \label{fig:subfig:b} 
    \includegraphics[width=0.8in]{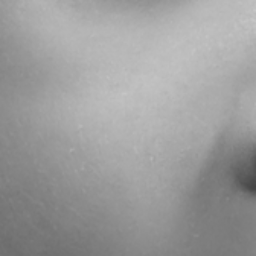}}
  \subfigure[Fake]{
    \label{fig:subfig:b} 
    \includegraphics[width=0.8in]{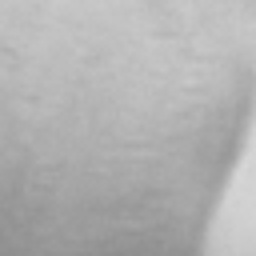}}
 \subfigure[Real]{
    \label{fig:subfig:b} 
    \includegraphics[width=0.8in]{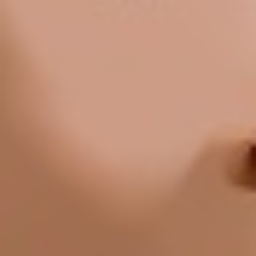}}
    \subfigure[Fake]{
    \label{fig:subfig:a} 
    \includegraphics[width=0.8in]{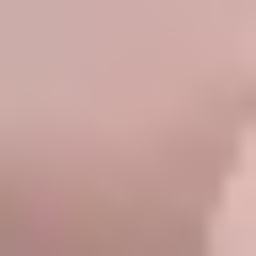}}
  \caption{Example images of Original (Skin) (a--b), Gray-scale (Skin) (c--d) and L0 filtered (Skin) (e--f).(better viewed in color)} 
  \label{fig:faceeye} 
\end{figure}

To validate the importance of textures for fake face detection, we conduct \textit{in-domain} experiments on the skin regions since they contain rich texture information and less structural information such as shape.
More specifically, we design the following controlled experiments on skin regions.
\begin{itemize}
\item \textit{Original (skin)}: The input is the left cheek skin region based on DLib \cite{dlib09} face alignment algorithm as shown in Figure~\ref{fig:faceeye} (a -- b). This is to verify whether the skin region contains enough useful information for fake face detection.
\item \textit{Gray-scale (skin)}: The skin regions are converted to gray-scale images. Typical examples are shown in  Figure~\ref{fig:faceeye} (c -- d). This experiment is to ablate the influence of color.
\item \textit{L0-filtered (skin)}: Small textures of the skin regions are filtered with $L_0$ filter~\cite{xu2011image}.
The $L_0$ algorithm can keep shape and color information while smoothing small textures. Typical examples are shown in Figure~\ref{fig:faceeye} (e -- f).
\end{itemize}
Experimental results are shown in Table \ref{tab:face} (row 3 -- row 5).
The results of full image, original skin region, gray-scale skin region as inputs all indicate that skin regions already contain enough information for \textit{in-domain} fake face detection and that colors do not influence the result much. 
The significant drop of performance (around $20\%$) of $L_0$ filtered inputs demonstrates the importance of texture for fake face detection in CNN models. In summary, texture plays a crucial role in CNN fake face detection and CNNs successfully capture the texture differences for discrimination, since the skin region performs on par with the full image in Table \ref{tab:face} (row 2 \& row 3).

\subsection{What are the differences between real \& fake faces in terms of texture?}\label{sec:texture-difference}
Empirical findings in Sec.~\ref{sec:texture-important} further motivate us to investigate the differences between real/fake faces in terms of {texture}.
In the following, we adopt a
{texture analysis} tool  -- the gray-level co-occurrence matrix (GLCM) \cite{haralick1973textural}.

The GLCM $P_\theta^d\in R^{256\times256}$ is created from a gray-scale {texture} image, and measures the co-occurrence of pixel values at a given offset parameterized by distance $d$ and angle $\theta$. For example, $P_{\theta}^d(i,j)$ indicates how often a pixel with value $i$ and a pixel at offset $(d, \theta)$ with pixel value $j$ co-exist.
In our analysis, we calculate $P_d^{\theta}$ across the whole dataset to get the statistical results, where $d\in\{1,2,5,10,15,20\}$ and $\theta\in \{0,\pi/2,\pi, 3\pi/2\}$  represents \{right, down, left, upper\},
$d$ and $\theta$ can capture the property of textures with different size and orientation respectively.
From the GLCM, we compute the texture contrast $\mathcal{C}_d$ at different distance offsets as follows,
\vspace{-0.1in}
\begin{center}
\begin{equation} \label{eq1}
\mathcal{C}_d  = \frac{1}{N}\sum_{i,j=0}^{255}\sum_{\theta=0}^{3\pi/2} |i-j|^2P_d^{\theta}(i,j)
\end{equation}
\end{center}
where $N = 256 \times 256 \times 4$ is a normalization factor, $i,j$ represents pixel intensities, and $d$ indicates pixel distances which
are adopted to compute $\mathcal{C}_d$. Larger $\mathcal{C}_d$ reflects stronger texture contrast, sharper and clearer visual effects. Inversely, low value $\mathcal{C}_d$ means the texture is blurred and unclear.

The contrast component of GLCM is shown in {Table~\ref{tab:texture1}}. Real faces retain \textit{stronger contrast} than fake faces at all measured distances. One explanation for this phenomenon is that the CNN based generator typically correlates the values of nearby pixels and cannot generate as strong texture contrast as real data.
In this section, we only provide an analysis of texture contrast and admit that the differences between real and fake faces are definitely beyond our analysis. We hope this can stimulate future research in analyzing the texture differences for fake face detection.

\section{Improved Model: Better Generalization Ability, More Robust}


Until now, our analysis has been performed in the \textit{in-domain} setting.
The next step is to investigate the \textit{cross-GAN} setting, where training and testing images are from different GAN models. Besides, we also investigate the images which are further modified by unintentional changes such as downsampling, JPEG compression and/or even intentional editing by adding blur or noise.
Our following analysis remains to focus on \textit{texture} due to our findings in Sec.~\ref{sec:human} -- Sec.~\ref{sec:texture-difference}.


\subsection{Generalization and Robustness Analysis}
\begin{table}
\centering
\scalebox{0.7}{
\begin{tabular}{c|c|c|c|c|c|c}
\toprule
\diagbox{Dataset}{distance ($d$)} & 1 & 2 & 5 & 10 &15 & 20   \\
\midrule
CelebA-HQ& \textbf{8.68} & \textbf{12.37} & \textbf{61.52} & \textbf{117.94} &\textbf{181.30}& \textbf{237.30}\\
StyleGAN(on CelebA-HQ)& 4.92 & 8.84 & 47.40 &93.79&146.33 & 193.49\\
PGGAN(on CelebA-HQ)&6.45 & 11.43 & 58.20 & 112.28 & 172.72  &226.40\\
 \bottomrule
\end{tabular}
}
\caption{Contrast property of GLCM calculated with all skin patches in training set.}
\label{tab:texture1}
\end{table}

\begin{figure*}[t]
  \centering
  \subfigure[Downsample]{
    \label{fig:subfig:a} 
    \includegraphics[height=1.5in]{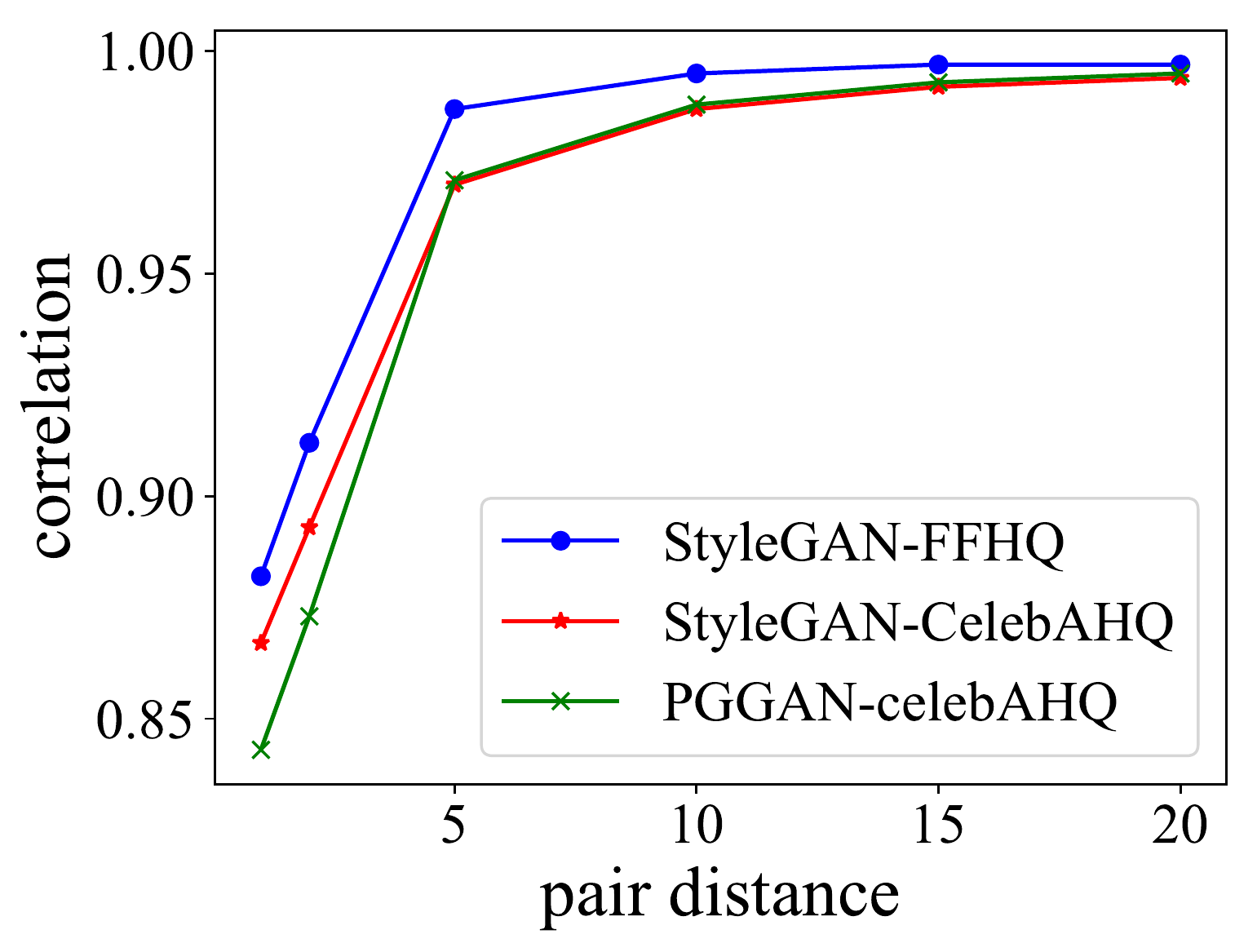}}
  \subfigure[Gaussian blur]{
    \label{fig:subfig:b} 
    \includegraphics[height=1.5in]{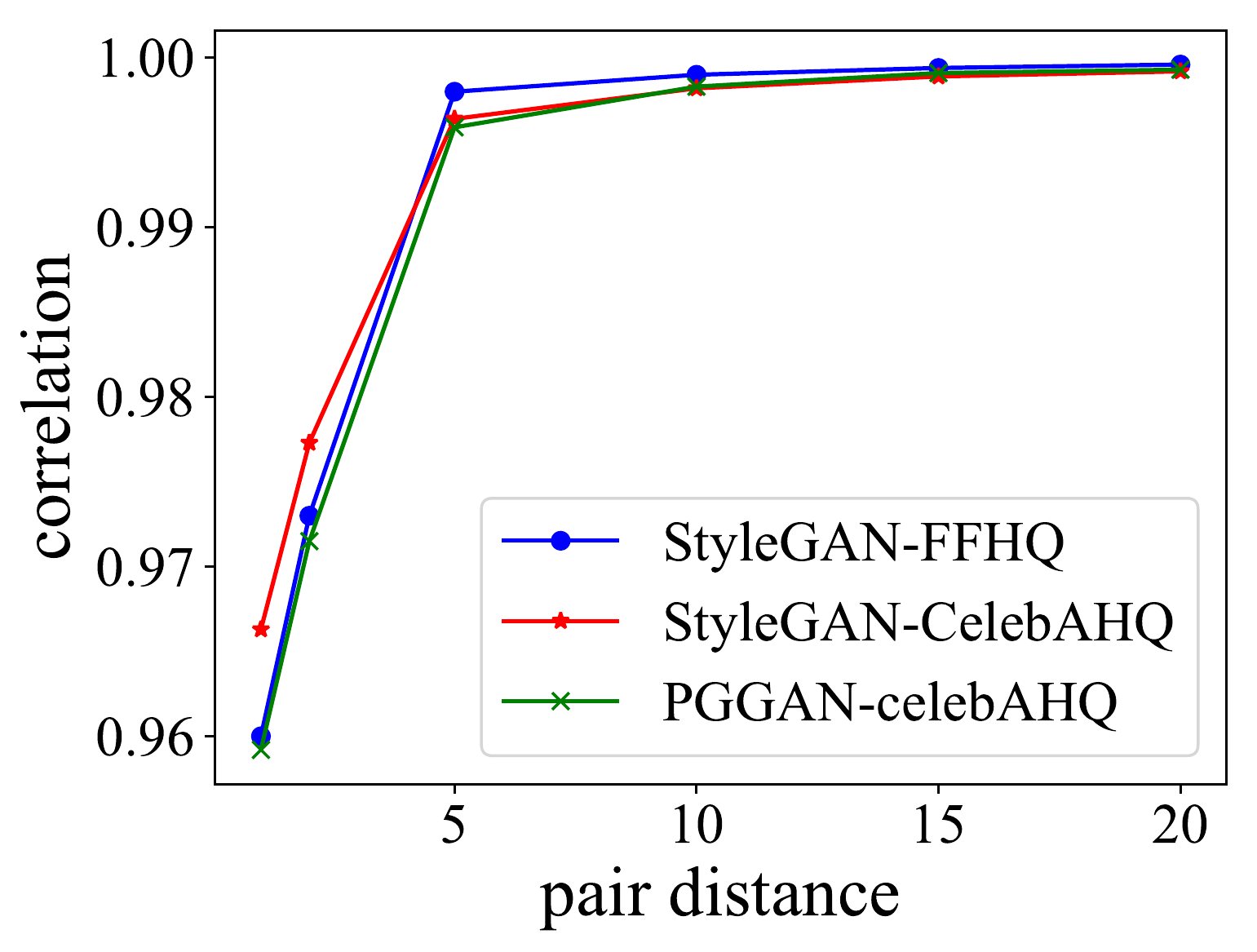}}
 \subfigure[Gaussian noise]{
 \label{fig:subfig:b} 
    \includegraphics[height=1.5in]{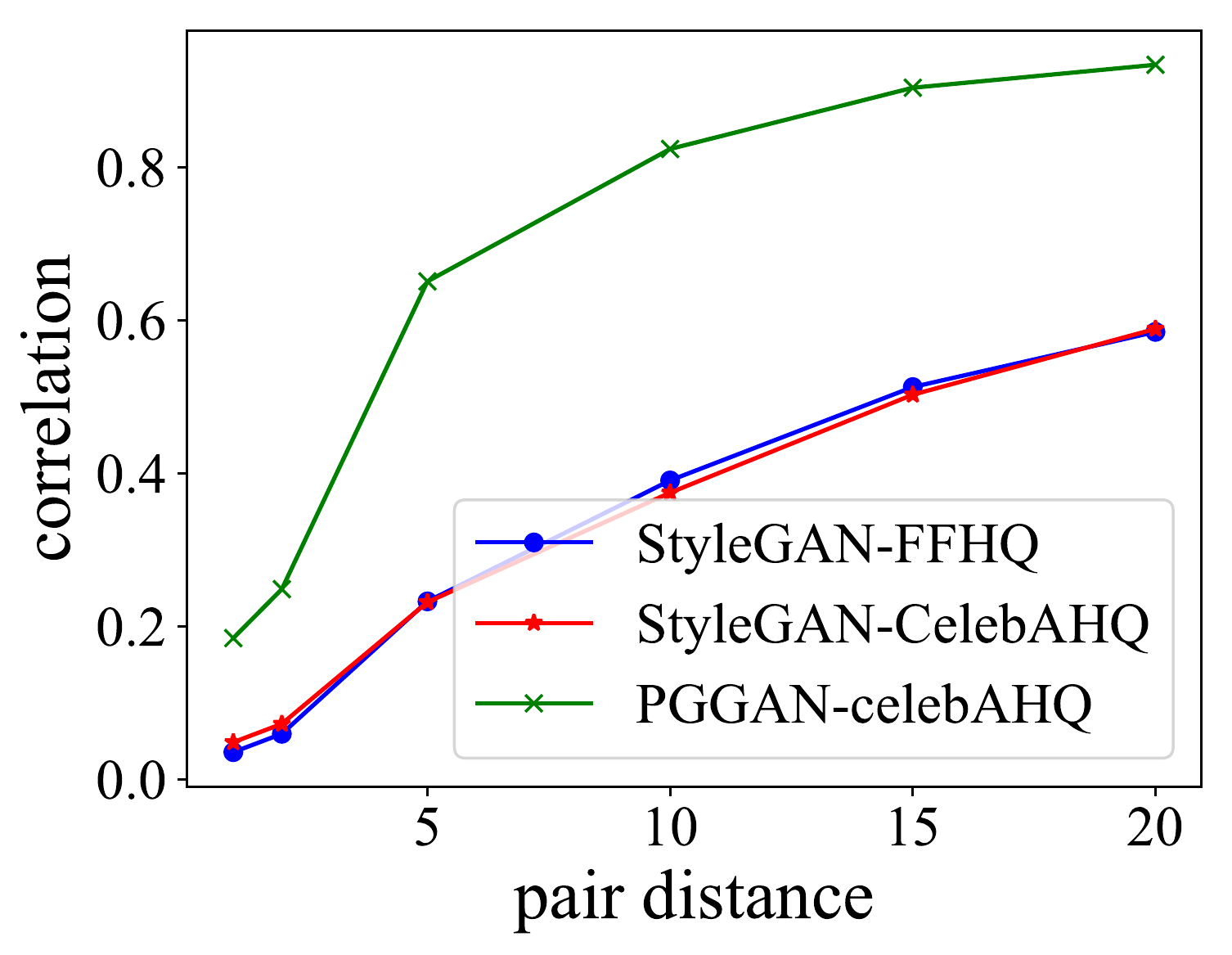}}
  \caption{Pearson correlation coefficient of texture contrast between edited images and original images.  Downsample ratio is $4$, Gaussian blur kernel is $3$, and  Guassian noise std is $3$.}
  \label{fig:correlation} 
\end{figure*}
Our previous experimental finding is that the trained model performs almost perfectly in \textit{in-domain} tests. However, our further experiments show that the performance of ResNet is reduced by $22\%$ (worst case) if the images are downsampled to $64 \times 64$ and JPEG compressed (Table \ref{tab:performance}: ``JPEG 8x $\downarrow$'').
Moreover, the model suffers more in \textit{cross-GAN} setting, especially when the trained models are evaluated on low-resolution GANs, in which the performance dropped to around $64\%-75\%$ (Table~\ref{tab:small}: Second row).
The reduction of performance indicates that the CNN fake/real image discriminator is not robust to image editing and cannot generalize well to \textit{cross-GAN} images, which limits its practical application.


To tackle the above problem, we further analyzed the issue.
In image editing scenario, we studied the correlation between the modified images and original ones.
Specifically, we calculate the Pearson Correlation Coefficient between the original image and edited images in terms of texture contrast $C_d$ as shown in Figure~\ref{fig:correlation}.
The coefficient value is closer to $1$ as the pair distance $d$ increases ({\ie} {larger image textures and more global}), which indicates a strong correlation in large texture between edited and original images.
In other words, large image texture has shown to be more robust to image editing.
Moreover, in \textit{cross-GAN} setting,
\textit{large} texture can also provide valuable information since the real/fake difference in terms of texture contrast  still hold in the large pair distance $d$ shown in Table ~\ref{tab:texture1}.
Thus a model that can capture long-range information is desirable to improve the model robustness and generalization ability.
However, current CNN models cannot incorporate long-range information due to its small effective receptive field which is much smaller than the calculated receptive field as presented in~\cite{luo2016understanding}.

\begin{figure*}[t]
\centering
\includegraphics[width=0.9\textwidth]{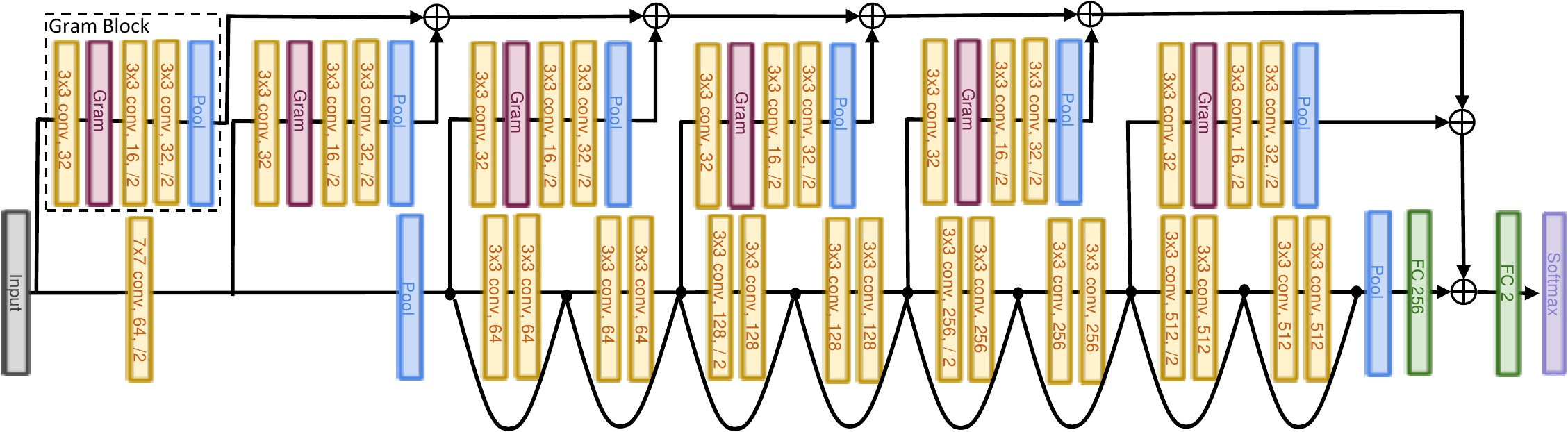}
\caption{Gram-Net architecture. We extract global image texture feature with 6 Gram Blocks in different semantic levels from ResNet. \textcircled{+} means concatenation.} \label{fig:gramnet}
\end{figure*}

Inspired by \cite{gatys2016image},
we propose to introduce ``Gram Block'' into the CNN architecture and propose a novel architecture coined as Gram-Net as shown in Figure~\ref{fig:gramnet}.
The ``Gram Block'' captures the global texture feature  and enable long-range modeling by calculating the Gram matrix in different semantic levels.


\begin{figure*}
  \centering
  \subfigure[Original]{
    \label{fig:subfig:a} 
    \includegraphics[width=1.0in]{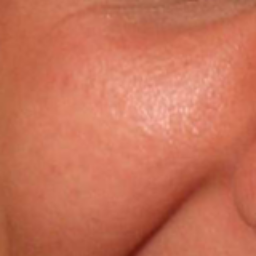}}
  \subfigure[ResNet]{
    \label{fig:subfig:b} 
    \includegraphics[width=1.0in]{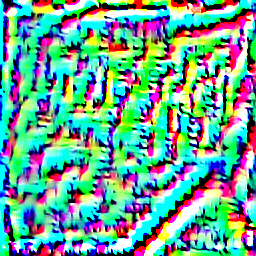}}
  \subfigure[Gram-Net]{
    \label{fig:subfig:b} 
    \includegraphics[width=1.0in]{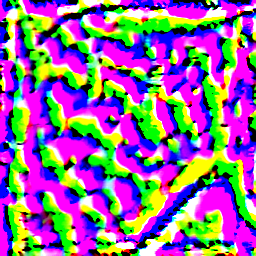}}
  \subfigure[ResNet]{
    \label{fig:subfig:b} 
    \includegraphics[width=1.0in]{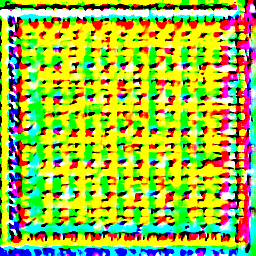}}
 \subfigure[Gram-Net]{
    \label{fig:subfig:b} 
    \includegraphics[width=1.0in]{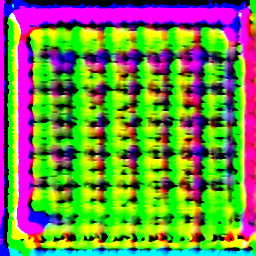}}
  \caption{Visualization of reconstructed input. Reconstructed images are multiplied by a scale factor for clearer visualization. (a) is the original image. (b)(c) are reconstructed inputs for reproducing 'res-block2' feature in ResNet and Gram-Net respectively. (d)(e) are reconstructed inputs for reproducing 'avg-pool' in ResNet and Gram-Net respectively.}      
  \label{fig:visualize} 
\end{figure*}

\subsection{Gram-Net Architecture}

The overview of Gram-Net is shown in Figure \ref{fig:gramnet}. Gram Blocks are added to the ResNet architecture on the input image and before every downsampling layer, incorporating global image texture information in different semantic levels. Each Gram Block contains a convolution layer to align the feature dimension from different levels, a Gram matrix calculation layer to extract global image texture feature, two conv-bn-relu layers to refine the representation, and a global-pooling layer to align the gram-style feature with ResNet backbone.
The Gram matrix is calculated as follows.

\begin{center}
\begin{equation} \label{eq2}
G_{ij}^l  = \sum_{k} F_{ik}^l F_{jk}^l
\end{equation}
\end{center}

where $F^l$ represents the $l$-th feature map whose spatial dimension is vectorized,
and $F_{ik}^l$ represents the $k$th element in the $i$th feature map of layer $l$. 
We show Gram matrix is a good descriptor for global or long-range texture as follows.

\paragraph{Can Gram matrix capture \textit{global texture} information?} In CNNs, each convolution layer $l$ can be viewed as a filter bank, and the feature map $F^l$ is a set of response images to these filters.

$G^l$ is the eccentric covariance matrix of channels in layer $l$.
Each element $G_{ij}^l$  measures the covariance between the $i$th and $j$th vectorised response map in the layer. Equation ~\ref{cov} is the covariance matrix $C^l$ of feature maps, and Gram matrix $G^l$ in Equation  ~\ref{gram} is actually the covariance matrix without subtracting the mean value.
The diagonal elements of Gram matrix shows the response of the particular filter, while other elements show the coherence of different filters. In a word, Gram matrix is a summary of spatial statistics which discards the spatial and content information in the feature maps, and provides a stationary description of the texture.

\begin{center}
\begin{footnotesize}
\begin{equation} \label{cov}
\begin{aligned}
C^l=(cov(F^l_i,F^l_j))_{n\times n}=(\mathop{\mathbb{E}}[(F^{lT}_i-\overline{F^{lT}_i})(F^l_j-\overline{F^l_j})])_{n\times n}= \\
\\
\frac{1}{n-1} \left[
   \begin{matrix}
  (F^{lT}_1-\overline{F^{lT}_1}) (F^l_1-\overline{F^l_1}) & \cdots & (F^{lT}_1-\overline{F^{lT}_1}) (F^l_n-\overline{F^l_n}) \\
  \vdots & \ddots &\\
  (F^{lT}_n-\overline{F^{lT}_n}) (F^l_1-\overline{F^l_1}) & \cdots  & (F^{lT}_n-\overline{F^{lT}_n}) (F^l_n-\overline{F^l_n}) \\
    \end{matrix}
  \right]
\end{aligned}
\end{equation}
\end{footnotesize}
\end{center}
\vspace{-0.25in}
\begin{center}
\begin{footnotesize}
\begin{equation} \label{gram}
G^l  =  (F_{i}^{lT} F^l_{j})_{n\times n} =  
  \left[
   \begin{matrix}
  F^{lT}_1 F^l_1 & \cdots & F^{lT}_1 F^l_n \\
  \vdots & \ddots &\\
  F^{lT}_n F^l_1 & \cdots  & F^{lT}_n F^l_n \\
    \end{matrix}
  \right]
\end{equation}
\end{footnotesize}
\end{center}

In addition, 
$G_{ij}^l$ is a descriptor for the whole feature map, which is not limited by the receptive field of CNNs.
This property enables it to extract long-range texture feature effectively, which complements the CNN backbone.

To further analyze the information captured by Gram-Net and the CNN baseline, we adopt~\cite{mahendran2015understanding} to generate the reconstructed input that can produce the approximate feature map as the original input.
The reconstructed inputs for reproducing the feature in ``res-block 2'' and ``avg-pool'' are shown in Figure~\ref{fig:visualize}.
The texture size of the reconstructed input image from Gram-Net is larger than that of baseline ResNet, which shows that our Gram-Net captures long-range texture patterns for discrimination.

\section{Experiments}
\begin{table*}[t]
\centering
\scalebox{0.8}{
\begin{tabular}{c|c|c|c|c|c|c|c|c|c}
\toprule
Training set & Testing set & Method & Original \% & 8x $\downarrow$ \% & JPEG \% &JPEG 8x $\downarrow$& Blur \% & Noise \% &Avg.  \\
\midrule

{} &{StyleGAN} & Co-detect &79.93 $\pm$ 1.34 & 71.80 $\pm$ 1.30 & 74.58 $\pm$ 3.25 & 71.25 $\pm $1.18& 71.39 $\pm $1.42& 54.09 $\pm$ 2.45&70.51\\
{StyleGAN} &{\vsss}& ResNet &  96.73 $\pm$ 3.60 & 85.10 $\pm$ 6.22 & 96.68 $\pm$ 3.50  &83.33 $\pm$ 5.95& 79.48 $\pm$ 8.70 & 87.92 $\pm$ 6.16 &88.20\\
{{\vsss}} &{CelebA-HQ} & Gram-Net &\textbf{99.10 $\pm$ 1.36}&\textbf{95.84 $\pm$ 1.98}&\textbf{99.05 $\pm$ 1.37}&\textbf{92.39 $\pm$ 2.66}&\textbf{94.20 $\pm$ 5.57} &\textbf{92.47 $\pm$ 4.52} &\textbf{95.51} \\

\cmidrule{2-10}
{CelebA-HQ}& {PGGAN} & Co-detect  & 71.22 $\pm$ 3.76 & 62.02 $\pm$ 2.86 &64.08 $\pm$ 1.93 & 61.24 $\pm$ 2.28&62.46 $\pm$ 3.31&49.96 $\pm$ 0.28&61.83\\
{}&{\vsss}& ResNet &93.74 $\pm$ 3.03  & 77.75 $\pm$ 4.82 & 89.35 $\pm$ 1.50 & 69.35 $\pm$ 3.25& 78.06 $\pm$ 7.57 & 82.65  $\pm$ 2.37&81.82\\
{} &{CelebA-HQ} & Gram-Net& \textbf{98.54} $\pm$ 1.27 & \textbf{82.40 $\pm$ 6.30}&\textbf{94.65 $\pm$ 3.28}&\textbf{79.77 $\pm$ 6.13}&\textbf{91.96 $\pm$ 4.78} & \textbf{88.29 $\pm$ 3.44} &\textbf{89.26}\\
 \midrule
 {} &{PGGAN} &
  Co-detect  &91.14 $\pm$ 0.61&82.94 $\pm$ 1.03& 86.00 $\pm$ 1.70 &82.46 $\pm$ 1.06&84.24 $\pm$ 0.93&54.77 $\pm$ 2.42&80.26\\
{PGGAN} &{\vsss} & ResNet &97.38 $\pm$ 0.52  & 90.87 $\pm$ 1.90 & 94.67 $\pm$ 1.15& 89.93 $\pm$ 1.50 & 97.25 $\pm$ 0.87 & 66.60 $\pm$ 9.61 &89.45\\
 {\vsss}&{CelebA-HQ} & Gram-Net &\textbf{98.78 $\pm$ 0.49}&\textbf{94.66 $\pm$ 3.10}&\textbf{97.29 $\pm$ 1.05}& \textbf{94.08 $\pm$ 3.22}&\textbf{98.55 $\pm$ 0.92}&\textbf{70.32 $\pm$ 12.04}&\textbf{92.28}\\
\cmidrule{2-10}
{CelebA-HQ}& {StyleGAN}&  Co-detect  &57.30 $\pm$ 1.62&57.41 $\pm$ 0.85&52.90 $\pm$ 1.67&82.46 $\pm$ 1.06&57.41 $\pm$ 0.93&50.08 $\pm$ 0.10&51.47\\
{} &{\vsss}& ResNet &97.98 $\pm$ 1.90 &87.91 $\pm$ 1.01 & 92.03 $\pm$ 4.14 & 82.23 $\pm$ 1.39 & 94.79 $\pm$ 1.32 & \textbf{60.89 $\pm$ 7.24}&85.97\\
{} &{CelebA-HQ} & Gram-Net &\textbf{98.55 $\pm$ 0.89}&\textbf{91.57 $\pm$ 2.95}&\textbf{94.28 $\pm$ 3.67}& \textbf{83.64 $\pm$ 3.43}&\textbf{97.05 $\pm$ 1.04}&60.07 $\pm$ 7.32&\textbf{87.52}\\
 \midrule

 {StyleGAN} & {StyleGAN} &Co-detect  &69.73 $\pm$ 2.41&67.27 $\pm$ 1.68&67.48 $\pm$ 2.83&64.65 $\pm$ 1.67&64.55 $\pm$ 1.93& 54.66  $\pm$ 3.97&64.74\\
{\vsss} &{\vsss} & ResNet &90.27 $\pm$ 3.05 & 70.99 $\pm$ 1.13 & 89.35 $\pm$ 3.42 &  67.96 $\pm$ 1.13& \textbf{75.60 $\pm$ 10.75}& 81.32 $\pm$ 5.06 &81.50\\
{FFHQ} &{FFHQ} & Gram-Net & \textbf{98.96 $\pm$ 0.51}& \textbf{89.22 $\pm$ 4.44}& \textbf{98.69 $\pm$ 0.81}&  \textbf{87.86 $\pm$ 3.42}&70.99 $\pm$ 6.07 &\textbf{94.27 $\pm$ 2.12} & \textbf{90.00}\\
 \bottomrule
\end{tabular}
}
\vspace{0.3cm}
\caption{Performance on in-domain and cross to high-resolution GANs. In each training setting, the first half shows results in the \textit{in-domain} setting and the second half shows results in the \textit{cross-GAN} setting.
Column (Avg.) shows the averaged results across all settings. The accuracy in ``Original \%'' column is lower than the results in Table \ref{tab:face} because the models are selected to achieve best average performance in all the settings with validation set. The average inference time for one image on RTX 2080 Ti are as follows. Gram-Net takes $2.40e^{-3}$s, ResNet-18 takes $2.35e^{-3}$s, and  Co-detect [27] takes $8.68e^{-3}$s, in which $6.57e^{-3}$s for co-occurance matrix calculation.}
\label{tab:performance}
\end{table*}

\paragraph{Implementation details.}
We implement all the approaches with PyTorch \cite{paszke2017automatic}.
Models are initialized with pretrained ImageNet weights.
We train all the models with learning rate $1e^{-5}$ and select model on validation set.
The validation set contains totally $800$ images from DCGAN, StarGAN, CelebA, PGGAN, StyleGAN on CelebA-HQ, StyleGAN on FFHQ, CelebA-HQ and FFHQ ($100$ for each).
In all the experiments, the models are trained on $10$k real and $10$k fake images and evaluated on a holdout test set containing $10$k real and $10$k fake images.
\vspace{-0.2cm}
\paragraph{Experimental setup.}
We conduct experiments in \textit{in-domain} and \textit{cross-GAN} settings, and further test the models on GANs trained on other datasets (\textit{cross-dataset}). All the images are bilinear-resized to  $512\times 512$ which serves our baseline resolution, because we found that models on this resolution already performs almost the same as $1024\times 1024$ and can accelerate the inference.
All fake images are derived by directly evaluating the author-released code and model with default parameters.
We compare the performance of our Gram-Net with a recent fake face detectors Co-detect ~\cite{nataraj2019detecting} and ResNet. We choose ResNet-18 as baseline because it already achieves much better performance than human beings described in Section \ref{sec:human}.
For a fair comparison, we implement Gram-Net and ~\cite{nataraj2019detecting} with the same ResNet-18 backbone, which takes the hand-craft texture descriptor GLCM of RGB channels as input. 
We train these three networks with images randomly bilinear-resized into range $64 \times 64$ to $256 \times 256$ as data augmentation, and evaluate the models regarding accuracy and their robustness to image editing and cross-GAN generalization ability.
To minimize the influence of randomness, we repeat each experiment five times by randomly splitting training and testing sets and show the error bar.

\vspace{-0.2cm}
\paragraph{Robustness and cross-GAN generalization experiments on high-resolution GANs.}\label{sec:high}
We edit the images with downsampling and JPEG compression, which often occur unintentionally when the images are uploaded to the Internet, put into slides or used as a video frame.
Specifically, the models are evaluated in the following settings. 1) Original inputs with size $512\times 512$ (``Origin''), 2) Downsampled images to resolution $64\times 64$ (``8x $\downarrow$''), 3) JPEG Compressed $512\times 512$ images (``JPEG''), 4) JPEG compressed and  downsampled images (``JPEG 8x $\downarrow$''). In addition, GAN and real images can be edited by adding blur or noise intentionally. 
In table \ref{tab:performance}, Gaussian blur (``blur'') is with kernel size $25$ (``blur''), and Gaussian noise (``blur'') is with standard deviation $5$.

The evaluation results are listed in Table \ref{tab:performance}.
Our Gram-Net outperforms the compared methods in all scenarios. On average, 
it outperforms \cite{nataraj2019detecting} by more than $20\%$. 
The results show that our Gram-Net adaptively extracts robust texture representation in feature space, which is much more powerful than low-level texture representations such as GLCM.
Our model also improves the ResNet baseline by around $7\%$ (on average) 
in both in-domain and cross-GAN settings trained on StyleGAN {\vsss} CelebA-HQ. The reason why Gram-Net improves less when trained on PGGAN {\vsss} CelebA-HQ can be partially explained according to the GLCM statistics shown in Table \ref{tab:texture1}. Images generated by PGGAN have larger  $\mathcal{C}_d$ than StyleGAN, which is closer to real images.


The above results manifest the effectiveness of Gram-Net in extracting features more invariant to different GAN models and more robust to image editing operations, such as downsampling, JPEG compression, blur and noise.

\begin{table*}
\centering
\scalebox{0.8}{
\begin{tabular}{c|c|c|c|c|c}
\toprule
\diagbox{Train}{Test} & Method & DCGAN {\vsss} CelebA \% & DRAGAN \ {\vsss} CelebA \% & StarGAN {\vsss} CelebA \% & Avg.  \\
\midrule
 {StyleGAN} & Co-detect &68.83 $\pm$ 9.57& 59.99 $\pm$ 8.81&58.60 $\pm$ 3.99 & 62.47\\
{\vsss} & ResNet &75.11 $\pm$ 8.10  & 65.53 $\pm$ 8.20 & 64.04 $\pm$ 7.69 & 68.22\\
 {CelebA-HQ} &  Gram-Net & \textbf{81.65 $\pm$ 3.51}&\textbf{76.40 $\pm$ 6.06}&\textbf{74.96 $\pm$ 4.90} & \textbf{77.67}\\
  \bottomrule
\end{tabular}
}
\vspace{0.3cm}
\caption{Performance of Gram-Net on generalization to low-resolution GANs.}
\label{tab:small}
\end{table*}

\vspace{-0.2cm}
\paragraph{Generalize to low-resolution GANs.}
To further evaluate the models' generalization capability, we directly apply the models above to low-resolution GANs trained on CelebA. 
We randomly choose $10$k images from each set to evaluate our model. The fake images are kept at their original sizes, {\ie}, 64$\times$64 for DCGAN and DRAGAN, 128$\times$128 for StarGAN. CelebA images are of size 178$\times$218, so we center crop the 178$\times$178 patch in the middle to make it square.

The results as listed in Table \ref{tab:small} show that our Gram-Net better generalizes to low-resolution GANs. 
The performance of baseline ResNet and \cite{nataraj2019detecting} degrades to around $50\%$ to $75\%$ in this setting.
However, our method outperforms the ResNet baseline by around $10\%$ and \cite{nataraj2019detecting} by around $15\%$ regarding accuracy in all settings.
This further demonstrates {global image texture} feature introduced by our ``Gram Block'' is more invariant  across different GANs, which can even generalize to detect fake faces from image-to-image translation model -- StarGAN.


\vspace{-0.1in}\begin{table}[H]
\centering
\scalebox{0.8}{
\begin{tabular}{c|c}
\toprule
Method & Accuracy\\
 \midrule
Co-detect & 59.81 $\pm$ 10.82 \\
ResNet & 80.55 $\pm$ 6.37 \\
Gram-Net & \textbf{93.35 $\pm$ 2.25} \\
 \bottomrule
\end{tabular}
}
\vspace{0.3cm}
\caption{Performance of Gram-Net when StyleGAN discriminator contains Gram-Block. The models are trained on StyleGAN (origin) {\vsss} CelebA-HQ and tested on StyleGAN (with Gram-Block in discriminator) {\vsss} CelebA-HQ.}
\label{tab:gramloss}
\end{table}

\begin{table*}
\centering
\scalebox{0.8}{
\begin{tabular}{c|c|c|c}
\toprule
\tabincell{c}{Method\\} & \tabincell{c}{Train on StyleGAN {\vsss} CelebA-HQ\\Test on StyleGAN {\vsss} FFHQ}  & \tabincell{c}{Train on PGGAN {\vsss} CelebA-HQ\\Test on StyleGAN {\vsss} FFHQ}& \tabincell{c}{Train on StyleGAN {\vsss} FFHQ\\Test on StyleGAN {\vsss} CelebA-HQ} \\
 \midrule
Co-detect & 48.90 $\pm$ 3.95 & 48.71 $\pm$ 1.43 & 59.22 $\pm$ 1.30\\
ResNet &  75.45 $\pm$ 7.01 & 54.44 $\pm$3.64 & 80.14 $\pm$ 7.47\\
Gram-Net & \textbf{77.69 $\pm$ 6.49} & \textbf{59.57 $\pm$ 8.07} & \textbf{80.72 $\pm$ 6.02}\\
 \bottomrule
\end{tabular}
}
\vspace{0.3cm}
\caption{Performance of Gram-Net in cross-dataset settings}
\label{tab:small2}
\end{table*}

\begin{table}
\centering
\scalebox{0.75}{
\begin{tabular}{c|c|c|c|c|c|c}
\toprule
{distance} & 1 & 2 & 5 & 10 &15 & 20   \\
\midrule
ImageNet & \textbf{525.70} & \textbf{676.60} & \textbf{1551.85} & \textbf{2267.16} &\textbf{2892.90}& \textbf{3334.14}\\
BigGAN & 367.65 & 536.81 & 1426.66 & 2146.90 & 2771.96 & 3207.97\\
 \bottomrule
\end{tabular}
}
\caption{Contrast property of GLCM calculated with BigGAN and ImageNet images in training set with different pair distances.}
\label{tab:texture_real}
\end{table}

\vspace{-0.3in}

\paragraph{Generalize to StyleGAN trained with Gram-Block in discriminator.}\label{sec:gramstylegan}
In this section, we evaluate the model on images from GAN models whose discriminator also contains Gram Blocks.
We fine-tune StyleGAN with extra Gram-Blocks inserted in the discriminator, and further evaluate whether Gram-Net still works in this setting. We add $8$ identical Gram-Blocks as in Gram-Net to encode feature maps (from feature map size 1024 to 4) in StyleGAN discriminator, and concatenate the 8$\times$32 dimension feature vector extracted by Gram-Blocks with the original 512 dimension feature vector in original discriminator before the final classification. We fine-tune the model for 8K epochs on CelebA-HQ initialized by the author released model.
We evaluate 10K fake images from StyleGAN with Gram-Block in discriminator and 10K images from CelebA-HQ. The images are resized to  $512 \times 512$ resolution. We directly apply the models used in Table \ref{tab:performance} and \ref{tab:small} in this setting.

The results in Table \ref{tab:gramloss} show that our Gram-Net still outperforms baseline methods even though the Gram-Block is inserted in the GAN discriminator. This demonstrates that our findings and analysis in section \ref{sec:texture-difference} are still valid.

\vspace{-0.2in}
\paragraph{Cross-dataset experiments.}\label{sec:cross-dataset}

Cross-dataset generalization  is a challenging problem due to the inherent difference in dataset construction.
Our experiments show that the statistics of CelebA-HQ and FFHQ are significantly different and can easily be distinguished by a neural network.
Specifically, we built a real face image dataset consisting of 10K CelebA-HQ images and 10K FFHQ images, and our further experiments show that a ResNet network can achieve more than 99.9\% accuracy to discriminate CelebA-HQ and FFHQ images. This experiment shows that real face datasets significantly differ from each other.

Despite the fact above, 
we evaluate our Gram-Net and baseline approaches in the cross-dataset setting as follows: train on StyleGAN(PGGAN) {\vsss} CelebA-HQ and test on StyleGAN {\vsss} FFHQ, train on StyleGAN {\vsss} FFHQ and test on StyleGAN {\vsss} CelebA-HQ. We keep all of the images as their original resolution in this experiment.
 The models are the same with the ones in Table \ref{tab:performance} and \ref{tab:small}.

The result in Table \ref{tab:small2} shows that fake image detectors trained on more realistic dataset (FFHQ) and stronger GANs (StyleGAN) have stronger ability to cross to less realistic datasets (CelebA-HQ) and less strong GANs (PGGAN). Also, Gram-Net still outperforms baselines methods.

\vspace{-0.2cm}
\paragraph{Generalize to natural images.}
In this section, we extend our analysis and apply Gram-Net to fake/real natural images. Specifically, we analyze ImageNet~\cite{krizhevsky2012imagenet} \vs BigGAN~\cite{brock2018large}, where the BigGAN model is trained on ImageNet.

To analyze fake/real natural images, we further employ GLCM. We find that the difference in terms of texture contract between fake and real face images also holds for natural images. As Table \ref{tab:texture_real} shows, real images retain stronger texture contrast than GAN images for all the distances measured.

To evaluate the generalization ability of our model trained on face images, we directly apply the model used in Table \ref{tab:performance} and \ref{tab:small} to test 10K ImageNet and 10K BigGAN images (10 images each class), and the results are shown in Table \ref{tab:real}. 

\begin{table}[H]
\centering
\scalebox{0.7}{
\begin{tabular}{c|c|c|c}
\toprule
Training set & Testing set & Method & Accuracy \% .  \\
\midrule
StyleGAN & ImageNet & Co-detect~\cite{nataraj2019detecting}  & 51.94 $\pm$ 2.31\\
\vs  &\vs & ResNet & 71.93 $\pm$ 2.09\\
CelebA-HQ & BigGAN  & Gram-Net & \textbf{80.29 $\pm$ 3.20}\\
 \bottomrule
\end{tabular}
}
\vspace{0.3cm}
\caption{Quantitative results on ImageNet vs BigGAN.}
\label{tab:real}
\end{table}

\section{Conclusion}

In this paper, we conduct empirical studies on human and CNNs in discriminating fake/real faces and find that fake faces attain different textures from the real ones. Then, we perform low-level texture statistical analysis to further verify our findings.
The statistics also show that \textit{large texture} information is more robust to image editing and invariant among different GANs.
Motivated by these findings, we propose a new architecture -- Gram-Net, which leverages \textit{global  texture} features to improve the robustness and generalization ability in fake face detection.
Experimental results show that Gram-Net significantly outperforms the most recent approaches and baseline models in all settings including \textit{in-domain}, \textit{cross-GAN}, and \textit{cross-dataset}.
Moreover, our model exhibits better generalization ability in detecting fake natural images.
Our work shows a new and promising direction for understanding fake images from GANs and improving fake face detection in the real world. 

\section{Acknowledgement}
This work was supported by the ERC grant ERC-2012-AdG 321162-HELIOS, EPSRC grant Seebibyte EP/M013774/1 and EPSRC/MURI grant EP/N019474/1. We would also like to acknowledge the Royal Academy of Engineering and FiveAI.

{\small
\bibliographystyle{ieee_fullname}
\bibliography{egbib}

\begin{thebibliography}{10}\itemsep=-1pt

\bibitem{arjovsky2017wasserstein}
Martin Arjovsky, Soumith Chintala, and L{\'e}on Bottou.
\newblock Wasserstein generative adversarial networks.
\newblock In {\em International Conference on Machine Learning}, pages
  214--223, 2017.

\bibitem{berthelot2017began}
David Berthelot, Thomas Schumm, and Luke Metz.
\newblock Began: Boundary equilibrium generative adversarial networks.
\newblock {\em arXiv preprint arXiv:1703.10717}, 2017.

\bibitem{brock2018large}
Andrew Brock, Jeff Donahue, and Karen Simonyan.
\newblock Large scale gan training for high fidelity natural image synthesis.
\newblock {\em arXiv preprint arXiv:1809.11096}, 2018.

\bibitem{choi2018stargan}
Yunjey Choi, Minje Choi, Munyoung Kim, Jung-Woo Ha, Sunghun Kim, and Jaegul
  Choo.
\newblock Stargan: Unified generative adversarial networks for multi-domain
  image-to-image translation.
\newblock In {\em Proceedings of the IEEE Conference on Computer Vision and
  Pattern Recognition}, pages 8789--8797, 2018.

\bibitem{gatys2015texture}
Leon Gatys, Alexander~S Ecker, and Matthias Bethge.
\newblock Texture synthesis using convolutional neural networks.
\newblock In {\em Advances in neural information processing systems}, pages
  262--270, 2015.

\bibitem{gatys2016image}
Leon~A Gatys, Alexander~S Ecker, and Matthias Bethge.
\newblock Image style transfer using convolutional neural networks.
\newblock In {\em Proceedings of the IEEE conference on computer vision and
  pattern recognition}, pages 2414--2423, 2016.

\bibitem{geirhos2018imagenet}
Robert Geirhos, Patricia Rubisch, Claudio Michaelis, Matthias Bethge, Felix~A
  Wichmann, and Wieland Brendel.
\newblock Imagenet-trained cnns are biased towards texture; increasing shape
  bias improves accuracy and robustness.
\newblock {\em arXiv preprint arXiv:1811.12231}, 2018.

\bibitem{goodfellow2014generative}
Ian Goodfellow, Jean Pouget-Abadie, Mehdi Mirza, Bing Xu, David Warde-Farley,
  Sherjil Ozair, Aaron Courville, and Yoshua Bengio.
\newblock Generative adversarial nets.
\newblock In {\em Advances in neural information processing systems}, pages
  2672--2680, 2014.

\bibitem{gulrajani2017improved}
Ishaan Gulrajani, Faruk Ahmed, Martin Arjovsky, Vincent Dumoulin, and Aaron~C
  Courville.
\newblock Improved training of wasserstein gans.
\newblock In {\em Advances in Neural Information Processing Systems}, pages
  5767--5777, 2017.

\bibitem{haralick1973textural}
Robert~M Haralick, Karthikeyan Shanmugam, et~al.
\newblock Textural features for image classification.
\newblock {\em IEEE Transactions on systems, man, and cybernetics},
  (6):610--621, 1973.

\bibitem{he2016deep}
Kaiming He, Xiangyu Zhang, Shaoqing Ren, and Jian Sun.
\newblock Deep residual learning for image recognition.
\newblock In {\em Proceedings of the IEEE conference on computer vision and
  pattern recognition}, pages 770--778, 2016.

\bibitem{karras2017progressive}
Tero Karras, Timo Aila, Samuli Laine, and Jaakko Lehtinen.
\newblock Progressive growing of gans for improved quality, stability, and
  variation.
\newblock {\em arXiv preprint arXiv:1710.10196}, 2017.

\bibitem{karras2018style}
Tero Karras, Samuli Laine, and Timo Aila.
\newblock A style-based generator architecture for generative adversarial
  networks.
\newblock {\em arXiv preprint arXiv:1812.04948}, 2018.

\bibitem{dlib09}
Davis~E. King.
\newblock Dlib-ml: A machine learning toolkit.
\newblock {\em Journal of Machine Learning Research}, 10:1755--1758, 2009.

\bibitem{kodali2017convergence}
Naveen Kodali, Jacob Abernethy, James Hays, and Zsolt Kira.
\newblock On convergence and stability of gans.
\newblock {\em arXiv preprint arXiv:1705.07215}, 2017.

\bibitem{krizhevsky2012imagenet}
Alex Krizhevsky, Ilya Sutskever, and Geoffrey~E Hinton.
\newblock Imagenet classification with deep convolutional neural networks.
\newblock In {\em Advances in neural information processing systems}, pages
  1097--1105, 2012.

\bibitem{li2018detection}
Haodong Li, Bin Li, Shunquan Tan, and Jiwu Huang.
\newblock Detection of deep network generated images using disparities in color
  components.
\newblock {\em arXiv preprint arXiv:1808.07276}, 2018.

\bibitem{li2018exposing}
Yuezun Li and Siwei Lyu.
\newblock Exposing deepfake videos by detecting face warping artifacts.
\newblock {\em arXiv preprint arXiv:1811.00656}, 2018.

\bibitem{liu2017unsupervised}
Ming-Yu Liu, Thomas Breuel, and Jan Kautz.
\newblock Unsupervised image-to-image translation networks.
\newblock In {\em Advances in Neural Information Processing Systems}, pages
  700--708, 2017.

\bibitem{liu2015faceattributes}
Ziwei Liu, Ping Luo, Xiaogang Wang, and Xiaoou Tang.
\newblock Deep learning face attributes in the wild.
\newblock In {\em Proceedings of International Conference on Computer Vision
  (ICCV)}, 2015.

\bibitem{luo2016understanding}
Wenjie Luo, Yujia Li, Raquel Urtasun, and Richard Zemel.
\newblock Understanding the effective receptive field in deep convolutional
  neural networks.
\newblock In {\em Advances in neural information processing systems}, pages
  4898--4906, 2016.

\bibitem{mahendran2015understanding}
Aravindh Mahendran and Andrea Vedaldi.
\newblock Understanding deep image representations by inverting them.
\newblock In {\em Proceedings of the IEEE conference on computer vision and
  pattern recognition}, pages 5188--5196, 2015.

\bibitem{marra2018detection}
Francesco Marra, Diego Gragnaniello, Davide Cozzolino, and Luisa Verdoliva.
\newblock Detection of gan-generated fake images over social networks.
\newblock In {\em 2018 IEEE Conference on Multimedia Information Processing and
  Retrieval (MIPR)}, pages 384--389. IEEE, 2018.

\bibitem{marra2019gans}
Francesco Marra, Diego Gragnaniello, Luisa Verdoliva, and Giovanni Poggi.
\newblock Do gans leave artificial fingerprints?
\newblock In {\em 2019 IEEE Conference on Multimedia Information Processing and
  Retrieval (MIPR)}, pages 506--511. IEEE, 2019.

\bibitem{marra2019incremental}
Francesco Marra, Cristiano Saltori, Giulia Boato, and Luisa Verdoliva.
\newblock Incremental learning for the detection and classification of
  gan-generated images.
\newblock {\em arXiv preprint arXiv:1910.01568}, 2019.

\bibitem{mccloskey2018detecting}
Scott McCloskey and Michael Albright.
\newblock Detecting gan-generated imagery using color cues.
\newblock {\em arXiv preprint arXiv:1812.08247}, 2018.

\bibitem{nataraj2019detecting}
Lakshmanan Nataraj, Tajuddin~Manhar Mohammed, BS Manjunath, Shivkumar
  Chandrasekaran, Arjuna Flenner, Jawadul~H Bappy, and Amit~K Roy-Chowdhury.
\newblock Detecting gan generated fake images using co-occurrence matrices.
\newblock {\em arXiv preprint arXiv:1903.06836}, 2019.

\bibitem{paszke2017automatic}
Adam Paszke, Sam Gross, Soumith Chintala, Gregory Chanan, Edward Yang, Zachary
  DeVito, Zeming Lin, Alban Desmaison, Luca Antiga, and Adam Lerer.
\newblock Automatic differentiation in pytorch.
\newblock 2017.

\bibitem{radford2015unsupervised}
Alec Radford, Luke Metz, and Soumith Chintala.
\newblock Unsupervised representation learning with deep convolutional
  generative adversarial networks.
\newblock {\em arXiv preprint arXiv:1511.06434}, 2015.

\bibitem{wang2019fakespotter}
Run Wang, Lei Ma, Felix Juefei-Xu, Xiaofei Xie, Jian Wang, and Yang Liu.
\newblock Fakespotter: A simple baseline for spotting ai-synthesized fake
  faces.
\newblock {\em arXiv preprint arXiv:1909.06122}, 2019.

\bibitem{xu2011image}
Li Xu, Cewu Lu, Yi Xu, and Jiaya Jia.
\newblock Image smoothing via l 0 gradient minimization.
\newblock In {\em ACM Transactions on Graphics (TOG)}, volume~30, page 174.
  ACM, 2011.

\bibitem{xuan2019generalization}
Xinsheng Xuan, Bo Peng, Jing Dong, and Wei Wang.
\newblock On the generalization of gan image forensics.
\newblock {\em arXiv preprint arXiv:1902.11153}, 2019.

\bibitem{yang2019exposing}
Xin Yang, Yuezun Li, Honggang Qi, and Siwei Lyu.
\newblock Exposing gan-synthesized faces using landmark locations.
\newblock {\em arXiv preprint arXiv:1904.00167}, 2019.

\bibitem{zhang2019detecting}
Xu Zhang, Svebor Karaman, and Shih-Fu Chang.
\newblock Detecting and simulating artifacts in gan fake images.
\newblock {\em arXiv preprint arXiv:1907.06515}, 2019.

\bibitem{zhou2016learning}
Bolei Zhou, Aditya Khosla, Agata Lapedriza, Aude Oliva, and Antonio Torralba.
\newblock Learning deep features for discriminative localization.
\newblock In {\em Proceedings of the IEEE conference on computer vision and
  pattern recognition}, pages 2921--2929, 2016.

\bibitem{zhu2017unpaired}
Jun-Yan Zhu, Taesung Park, Phillip Isola, and Alexei~A Efros.
\newblock Unpaired image-to-image translation using cycle-consistent
  adversarial networks.
\newblock In {\em Proceedings of the IEEE international conference on computer
  vision}, pages 2223--2232, 2017.

\end{thebibliography}
}

\end{document}